%% file: main.tex
\pdfoutput=1

\documentclass[11pt,a4paper]{article}
\usepackage[hyperref]{emnlp_files/emnlp2020}
\usepackage{times}
\usepackage{latexsym}
\usepackage{graphicx}
\usepackage{subfig}
\usepackage{epsfig}
\usepackage{xspace}
\usepackage{booktabs} 
\usepackage{array}
\usepackage{bm} 
\usepackage{sidecap} 
\usepackage{multirow}

\usepackage{microtype}

\aclfinalcopy 
\def\aclpaperid{404} 

\newcommand{\squeezeup}{\vspace{-4mm}}
\newcommand{\languageShort}[1]{#1\xspace}
\newcommand{\frenchShort}[0]{\languageShort{FR}}
\newcommand{\germanShort}[0]{\languageShort{DE}}
\newcommand{\italianShort}[0]{\languageShort{IT}}
\newcommand{\spanishShort}[0]{\languageShort{ES}}
\newcommand{\dutchShort}[0]{\languageShort{NL}}
\newcommand{\tableRef}[1]{Table \ref{#1}}
\newcommand{\sectionRef}[1]{\S \ref{#1}}
\newcommand{\figureRef}[1]{Figure \ref{#1}}
\newcommand{\argumentDatasetName}[0]{ArgsEN\xspace} 
\newcommand{\evidenceDatasetName}[0]{EviEN\xspace}
\newcommand{\humanGeneratedArgDatasetName}[0]{ArgsHG\xspace} 
\newcommand{\pseudoTestArgDatasetName}[0]{ArgsMT\xspace} 
\newcommand{\pseudoTestEviDatasetName}[0]{EviMT\xspace}

\newcommand{\datasetURL}[0]{\footnote{\url{https://www.research.ibm.com/haifa/dept/vst/debating_data.shtml\#Multilingual Argument Mining}}}
\newcommand{\DL}[0]{\emph{DL}\xspace} 
\newcommand{\RL}[0]{\emph{RL}\xspace}
\newcommand{\TL}[0]{\emph{TL}\xspace} 
\newcommand{\sixL}[0]{\emph{6L}\xspace}

\newcommand{\figureEightName}[0]{Appen\xspace}
\newcommand{\agreementName}[0]{agreement-$\kappa$\xspace}
\newcommand{\supp}[0]
{Appendix\xspace}
\newcommand{\forFinal}[1]{}

\newcolumntype{E}{@{\extracolsep{0.8cm}}c@{\extracolsep{0pt}}}

\newtheorem{theorem}{Theorem}
\newtheorem{example}[theorem]{Example}

\input{tables/table_column_commands}
\input{generated_commands}

\title{Multilingual Argument Mining: Datasets and Analysis}

\author{Orith Toledo-Ronen, Matan Orbach, Yonatan Bilu, Artem Spector and Noam Slonim\\
  IBM Research\\
  \texttt{\{oritht, matano, yonatanb, artems, noams\}@il.ibm.com}\\}
  
\date{}

\begin{document}
\maketitle
\input{abstract}
\input{introduction}
\input{related_work}

\input{data_sets}

\input{evaluation}

\input{results}

\input{conclusions}

\section*{Acknowledgments}
Special thanks go to Michal Jacovi, Christopher Sciacca, 
and Mohamed Nasr for coordinating the manual translation effort. 
We thank the teams who performed the translation, and Benjamin Sznajder for helping with the analysis.
We also thank the anonymous reviewers for their valuable comments.

\bibliography{main}
\bibliographystyle{emnlp_files/acl_natbib}

\input{Appendix/appendix}

\end{document}

%% file: tables/table_column_commands.tex
\newcommand{\numLabeledColumn}[0]{\#L}
\newcommand{\kappaColumn}[0]{$\bm{\kappa}$}
\newcommand{\supportingArgumentsColumn}[0]{Sup.}
\newcommand{\highQualityArgumentsColumn}[0]{HQ}
\newcommand{\languageColumn}[0]{Language}
\newcommand{\argumentCountColumn}[0]{\#C}

\newcommand{\argumentStanceCorrelationColumn}[0]{Stance-A}
\newcommand{\qualityCorrelationColumn}[0]{Qual}
\newcommand{\evidenceCorrelationColumn}[0]{Det}
\newcommand{\evidenceStanceCorrelationColumn}[0]{Stance-E}

\newcommand{\machineTranslated}[0]{MT\xspace}
\newcommand{\humanTranslated}[0]{HT\xspace}
\newcommand{\taskColumn}[0]{}

%% file: generated_commands.tex
\newcommand{\percentage}[1]{$#1$\%\xspace}
\newcommand{\statistic}[1]{$#1$\xspace}

\newcommand{\frHumanGeneratedStanceKappaUnfiltered}[0]{\statistic{0.66}}
\newcommand{\frHumanGeneratedStancePriorUnfiltered}[0]{\statistic{0.49}}

\newcommand{\frLabeledArgumentsUnfiltered}[0]{\statistic{1109}}
\newcommand{\deHumanGeneratedStanceKappaUnfiltered}[0]{\statistic{0.60}}
\newcommand{\deHumanGeneratedStancePriorUnfiltered}[0]{\statistic{0.50}}

\newcommand{\deLabeledArgumentsUnfiltered}[0]{\statistic{801}}
\newcommand{\itHumanGeneratedStanceKappaUnfiltered}[0]{\statistic{0.82}}
\newcommand{\itHumanGeneratedStancePriorUnfiltered}[0]{\statistic{0.50}}

\newcommand{\itLabeledArgumentsUnfiltered}[0]{\statistic{987}}
\newcommand{\esHumanGeneratedStanceKappaUnfiltered}[0]{\statistic{0.73}}
\newcommand{\esHumanGeneratedStancePriorUnfiltered}[0]{\statistic{0.51}}

\newcommand{\nlHumanGeneratedStanceKappaUnfiltered}[0]{\statistic{0.72}}
\newcommand{\nlHumanGeneratedStancePriorUnfiltered}[0]{\statistic{0.47}}

\newcommand{\nlLabeledArgumentsUnfiltered}[0]{\statistic{599}}

\newcommand{\frHumanGeneratedQualityKappaFiltered}[0]{\statistic{0.41}}
\newcommand{\frHumanGeneratedQualityPriorFiltered}[0]{\statistic{0.53}}
\newcommand{\frLabeledArgumentsFiltered}[0]{\statistic{903}}

\newcommand{\deHumanGeneratedQualityKappaFiltered}[0]{\statistic{0.39}}
\newcommand{\deHumanGeneratedQualityPriorFiltered}[0]{\statistic{0.56}}
\newcommand{\deLabeledArgumentsFiltered}[0]{\statistic{801}}

\newcommand{\itHumanGeneratedQualityKappaFiltered}[0]{\statistic{0.24}}
\newcommand{\itHumanGeneratedQualityPriorFiltered}[0]{\statistic{0.67}}
\newcommand{\itLabeledArgumentsFiltered}[0]{\statistic{969}}

\newcommand{\esHumanGeneratedQualityKappaFiltered}[0]{\statistic{0.29}}
\newcommand{\esHumanGeneratedQualityPriorFiltered}[0]{\statistic{0.62}}
\newcommand{\esLabeledArgumentsFiltered}[0]{\statistic{828}}

\newcommand{\nlHumanGeneratedQualityKappaFiltered}[0]{\statistic{0.40}}
\newcommand{\nlHumanGeneratedQualityPriorFiltered}[0]{\statistic{0.49}}
\newcommand{\nlLabeledArgumentsFiltered}[0]{\statistic{382}}

\newcommand{\esNumUniqueCollectedArguments}[0]{\statistic{2995}}
\newcommand{\frNumUniqueCollectedArguments}[0]{\statistic{2201}}
\newcommand{\deNumUniqueCollectedArguments}[0]{\statistic{1962}}
\newcommand{\itNumUniqueCollectedArguments}[0]{\statistic{3018}}
\newcommand{\nlNumUniqueCollectedArguments}[0]{\statistic{925}}

\newcommand{\deHTQualityStrongAllPearsonCorrelation}[0]{\statistic{0.49}}

\newcommand{\deHTStancePearsonCorrelation}[0]{\statistic{0.98}}

\newcommand{\deMTQualityStrongAllPearsonCorrelation}[0]{\statistic{0.45}}

\newcommand{\deMTStancePearsonCorrelation}[0]{\statistic{0.96}}

\newcommand{\deTranslatedQualityKappaFiltered}[0]{\statistic{0.26}}

\newcommand{\deTranslatedStanceKappaFiltered}[0]{\statistic{0.74}}

\newcommand{\itHTQualityStrongAllPearsonCorrelation}[0]{\statistic{0.47}}

\newcommand{\itHTStancePearsonCorrelation}[0]{\statistic{0.98}}

\newcommand{\itMTQualityStrongAllPearsonCorrelation}[0]{\statistic{0.32}}

\newcommand{\itMTStancePearsonCorrelation}[0]{\statistic{0.95}}

\newcommand{\itTranslatedQualityKappaFiltered}[0]{\statistic{0.12}}

\newcommand{\itTranslatedStanceKappaFiltered}[0]{\statistic{0.82}}

\newcommand{\deEvidenceDetectionKappa}[0]{\statistic{0.38}}

\newcommand{\deEvidenceStanceKappa}[0]{\statistic{0.82}}

\newcommand{\deHTEvidenceStanceSixLabelsPearsonCorrelation}[0]{\statistic{0.88}}

\newcommand{\deHTStrongEvidenceDetectionPearsonCorrelation}[0]{\statistic{0.75}}

\newcommand{\deMTEvidenceStanceSixLabelsPearsonCorrelation}[0]{\statistic{0.86}}

\newcommand{\deMTStrongEvidenceDetectionPearsonCorrelation}[0]{\statistic{0.74}}
\newcommand{\itEvidenceDetectionKappa}[0]{\statistic{0.33}}

\newcommand{\itEvidenceStanceKappa}[0]{\statistic{0.76}}

\newcommand{\itHTEvidenceStanceSixLabelsPearsonCorrelation}[0]{\statistic{0.80}}

\newcommand{\itHTStrongEvidenceDetectionPearsonCorrelation}[0]{\statistic{0.81}}

\newcommand{\itMTEvidenceStanceSixLabelsPearsonCorrelation}[0]{\statistic{0.85}}

\newcommand{\itMTStrongEvidenceDetectionPearsonCorrelation}[0]{\statistic{0.84}}

%% file: abstract.tex
\begin{abstract}
The growing interest in argument mining and computational argumentation brings with it a plethora of Natural Language Understanding (NLU) tasks and corresponding datasets. 
However, as with many other NLU tasks, the dominant language is English, with resources in other languages being few and far between.
In this work, we explore the potential of transfer learning using the multilingual BERT model to address argument mining tasks in non-English languages, based on English datasets and the use of machine translation. 
We show that such methods are well suited for classifying the stance of arguments and detecting evidence, but less so for assessing the quality of arguments, presumably because quality is harder to preserve under translation.
In addition, focusing on the translate-train approach, we show how the choice of languages for translation, and the relations among them, affect the accuracy of the resultant model. 
Finally, to facilitate evaluation of transfer learning on argument mining tasks, we provide a human-generated dataset with more than $10$k arguments in multiple languages, as well as machine translation of the English datasets.
\end{abstract}

%% file: introduction.tex
\section{Introduction}
\label{sec:intro}

Argument mining has received much attention in recent years, with research mainly focused on English and, to some extent, German texts. Recent advancements in Natural Language Understanding suggest that in order to train appropriate models for argument mining tasks in other languages, we do not need to manually label text in these languages, but rather employ transfer learning from the English-based models \cite{eger-etal-2018-cross}.

In this work we examine three argument mining tasks: (1) \emph{stance classification}: given a topic and
an argument that supports or contests the topic, determine the argument's stance towards the topic; (2) \emph{evidence detection}: given a topic and a sentence, determine if the sentence is an evidence relevant to the topic; (3) \emph{argument quality}: given a topic and a relevant argument, rate the argument so that higher-quality arguments are assigned a higher score.

To facilitate transfer learning from English datasets for these tasks, we employ Multilingual BERT (mBERT) released by \citet{devlin2019bert}, 
a pre-trained language model that supports 104 languages, and use it mainly in a translate-train approach. 
Namely, the English dataset is automatically translated into the desired language(s) using machine translation (MT); an augmented dataset composed of the original English text and all the translated copies is created; the mBERT model is fine-tuned on a subset of the dataset; and the resultant model is then used to solve the relevant downstream task in the desired language. 
Previous works have suggested that translating the original dataset to as large a number of languages as possible is beneficial \cite{liang2020xglue}. 
In this work, we show a more nuanced picture, where often selecting a subset of related languages is preferable.

In addition, we also examine the translate-test approach, in which one creates a classification model only with English data.
At prediction time, the non-English text is automatically translated into English, and then analyzed by the model.  
This approach is less appealing since the initial translation step increases prediction run-time, and on our data also tends to perform worse.

We examine two text sources for performance evaluation on non-English texts. The first one is a "pseudo test" set -- an automatic translation of an English evaluation set for a task. 
While such texts can be easily generated, it is not clear how well they represent "real" texts, authored by humans. Hence, we also examine human-authored texts, in several non-English languages, collected via crowdsourcing specifically for this work. 
Both datasets are released as part of this work.\datasetURL

When translating the evaluation set, either automatically or by a human translator, one would like to assume that the initial label of the English text is maintained after translation.
While this is often the case, we show that this assumption becomes more dubious as the argument mining task becomes more complex and subjective, as well as when the original labels are not clearly agreed upon.

In summary, the main contributions of this paper are:
(1) a comparative analysis of the translate-train approach on three central argument mining tasks using different subsets of languages,
showing that training on more data helps, but that, in some cases, training on related languages is sufficient; 
(2) multilingual benchmark datasets for the three tasks; 
(3) an analysis of the three tasks, showing how well labels are preserved across translation, and the impact that has on the success of the translate-train approach.

%% file: related_work.tex
\section{Related Work}
\label{sec:related_work}
Argument mining has been expanding from identifying argumentative passages to a variety of Natural Language Understanding (NLU) tasks (\citet{stede2018argumentation,lawrence2020argument}). 
In this work, we explore three argumentation tasks in multilingual settings.

Stance detection (or classification) is often contrasted with sentiment analysis, in that the task is not simply to classify the sentiment of a text, but rather its stance w.r.t. some given target. Early work on this task includes \citet{thomas2006get, lin2006side}, while in the context of argument mining, the task has probably been introduced by \citet{sobhani2015argumentation}. 
As with many other NLU tasks, earlier works developed classifiers based on various features (e.g., \citet{barhaim2017stance}), while more modern approaches rely on deep learning. See \citet{schiller2020stance} for a recent benchmarking report on such methods. 

Research on stance detection in a multilingual setting is rather recent. \citet{zotova2020multilingual} explore stance detection in Twitter for Catalan and Spanish; \citet{lai2020multilingual} do this for political debates in social media in these two languages as well as French and Italian; \citet{vamvas2020x} analyze the stance of comments in the context of the Switzerland election in German, French and Italian. 
Stance detection is reminiscent of the Natural Language Inference (NLI) problem, where one is given two sentences, and the objective is to determine whether one entails the other, contradicts it or is neutral. This task had been researched extensively, with \citet{conneau2018xnli} providing a 15-language benchmark for the multilingual setting. Another earlier work on a related task is addressing the support/attack relation prediction of two arguments in Italian \cite{basile-2016-arg-italian}.

Evidence detection is the task of determining, given some text and a topic, whether the text can serve as evidence in the context of the topic \cite{rinott-etal-2015-show}. 
We follow \citet{working-solution}, in defining \emph{evidence} as \textit{a single sentence that clearly supports or contests the 
topic, yet is not merely a belief or a claim. 
Rather, it provides an indication for whether a relevant belief or a claim is true}, and, since we use their datasets, we restrict our analysis to evidence of type Study and Expert. 
In a multilingual setting, a similar task, that of premise detection, was considered in \citet{eger-etal-2018-cross} for German, French, Spanish, and Chinese; in \citet{fishcheva2019cross} for Russian; in \citet{eger2018pd3} for French and German; and in \citet{aker2017projection} for Chinese. 

Argument quality prediction is the task of evaluating the quality of an argument, either on an objective scale - the input is an argument and the output is a quality score; or in a relative manner - the input is a pair of arguments and the output is which of them is of higher quality. 
While there are many, arguably independent, dimensions for quality \cite{wachsmuth2017computational}, it seems that people - and, consequently, algorithms - can usually perform this task in a consistent manner \cite{habernal2016argument, wachsmuth-etal-2017-argumentation, toledo-etal-2019-quality,gretz2020quality}. 
To the best of our knowledge, this task was not previously considered in a multilingual setting.

In contrast with previous multilingual research on argument mining, in this work we address three different problems, of varying complexity, over a relatively large number of languages. 
This allows us to draw more holistic conclusions on the efficacy -- and pitfalls -- of transfer learning in the argument mining domain. 
It is interesting to compare these conclusions with other wide-scope multilingual NLU research, such as the XTREME \cite{hu2020xtreme} and XGLUE \cite{liang2020xglue}.

%% file: data_sets.tex
\section{Data Sets}
\label{sec:datasets}

\subsection{Translated Data}
\label{par:argmining_data}
\paragraph{English Datasets}
The sources for our translated training data and the ”pseudo test” sets are two existing argument mining datasets in English, collected by our colleagues as part of our work on Project Debater.\footnote{Project Debater is the first AI system that can debate humans on complex topics: \url{https://www.research.ibm.com/artificial-intelligence/project-debater/}} 
One is a corpus of 30,497 arguments on $71$ controversial topics, annotated for their stance towards the topic and for their quality \cite{gretz2020quality}. This dataset (referred herein as \argumentDatasetName) is used for the stance classification and argument quality tasks. 

The second dataset is a corpus of 35,211 sentences from Wikipedia on 321 controversial topics, annotated for their stance towards the topic and the extent to which they can serve as an evidence for the topic \cite{working-solution}.
This dataset (referred as \evidenceDatasetName) is used for the stance classification and evidence detection tasks. 
Example \ref{ex:data} shows an argument and an evidence for one topic.

\input{data_examples}

A third dataset was used to augment the training data for evidence detection and stance classification -- the so-called VLD dataset of \citet{working-solution}, which includes around 200k sentences from newspaper articles,  
pertaining to 337 topics.\footnote{The stance labels of the \evidenceDatasetName and VLD datasets are not part of their official releases. They are included in our release.}

\paragraph{Data Selection}
\label{subsec:data_selection}
The \argumentDatasetName dataset was filtered for stance classification by selecting arguments with a clear stance (confidence $> 0.75$) for training and evaluation. 
For argument quality, arguments with a clear agreement on their quality were selected -- quality score above $0.9$ or below $0.4$. 

A positive label for evidence detection was assigned for evidence from \evidenceDatasetName with a score above $0.7$, and a negative label to those with score below $0.3$ (those with in-between scores were not used).
For stance classification on the evidence data, all sentences with a non-neutral stance were selected, since the \evidenceDatasetName dataset does not provide a confidence score for the stance label.

The VLD corpus was also filtered, taking sentences with evidence score above $0.95$ or below $0.05$.
This yielded a total of 52,037 sentences for training, of which
19,406 have a positive label. 

\paragraph{Translation}
\label{par:mt_data}
We used the Watson Language Translator\footnote{\url{www.ibm.com/watson/services/language-translator/}} to translate the selected English data into $5$ languages: German (DE), Dutch (NL), Spanish (ES), French (FR), and Italian (IT).
The translation 
is a one-time process, which can be applied to any target language (TL) that the MT engine in use supports.
The labels for the MT data were projected from the English data.

Following the data splits provided in the official release of each English corpus (into a training, development, and test sets), the translations of the training data were used for fine-tuning mBERT, and the translations of the test data were used for evaluation.
The translations of the test data of \argumentDatasetName and \evidenceDatasetName\xspace into the $5$ non-English languages -- the pseudo test sets -- are herein referred to as \pseudoTestArgDatasetName and \pseudoTestEviDatasetName.
The statistics of the translated English data are summarized in \tableRef{tab:selected_english_data_stats}.

\subsection{Human-Generated Data}
\label{par:hg_data}

\input{tables/selected_english_data_stats}

Arguments written in the TL provide a more realistic evaluation set than translated texts, specifically for tasks where labels are not well-preserved across automatic translation. 
Therefore, we created a new multilingual evaluation set by collecting arguments in all $5$ languages (ES, FR, IT, DE, and NL) 
for all the $15$ topics of the  \argumentDatasetName test set, using the \textit{\figureEightName}\footnote{\url{www.appen.com}} crowdsourcing platform. 
The human-authored evaluation dataset is herein referred to as \humanGeneratedArgDatasetName.

\paragraph{Annotation Setup}
Initially, crowd contributors wrote up to two pairs of arguments per topic, with one argument supporting the topic and another contesting it in each pair.
Next, the arguments were assessed for their stance and quality by $10$ annotators (per-language).
Given an argument, they were asked to determine the stance of the argument towards the topic and to assess whether it is of high quality.
The full argument annotation guidelines are included in the \supp, and \tableRef{tab:human_generated_labeling_stats} details the number of arguments collected and labeled per language.
To set a common standard, annotators were instructed to mark about half of the arguments they labeled as high quality. 
Annotation quality was controlled by integrating test questions (TQs) with an a-priori known answer in between the regular questions, measuring the per-annotator accuracy on these questions, and excluding underperformers.

A per-annotator average agreement score was computed by considering all peers sharing at least $50$ common answers, calculating pairwise Cohen's Kappa \cite{cohenKappa} with each of them, and averaging. 
Those not having at least $5$ peers meeting this criterion  
were excluded and their answers were discarded. 
Averaging the annotator agreements yields the average inter-annotator agreement (\agreementName) of each question.

To derive a label (or score) for each question we use the WA-score of \citet{gretz2020quality}. 
Roughly, answers are aggregated with a weight proportional to the agreement score for the annotators who chose them. 
At least $5$ answers were required for a question to be considered as labeled. 

Scaling the annotation from English to new languages required some adjustments, such as restricting participation to countries in which the TL is commonly spoken, and the use of TQs for the argument quality question. Further details are provided in the \supp.

\input{tables/human_generated_corpus}

\paragraph{Results} 
\tableRef{tab:human_generated_labeling_stats} presents the \agreementName for all TLs and each task for the human-generated dataset. 
For stance, the agreement is comparable to previously reported values for English ($0.69$ by \citet{toledo-etal-2019-quality} and $0.83$ for \argumentDatasetName).
For quality, the agreement is significantly better than previously reported on \argumentDatasetName ($0.12$ by \citet{gretz2020quality}),  presumably due to the use of TQs in this task, which were not included before.
The annotation in each of the non-English languages involved a distinct group of annotators, producing varying annotation quality among languages which is reflected in their \agreementName values.

The results also include the percentage of arguments labeled as supporting arguments, computed separately for each annotator and averaged over all annotators.
All values are close to $0.5$, confirming that the collected arguments are balanced for stance, as instructed.
Similarly, the results show the  percentage of arguments labeled as high quality, averaged over all annotators, confirming that annotators mostly followed the instruction to label about half of the arguments as high quality.

The same confidence filtering thresholds described in \sectionRef{subsec:data_selection} were applied to 
the data of \humanGeneratedArgDatasetName.
The statistics of the arguments selected for evaluation are shown in \tableRef{tab:human_generated_labeling_stats} (right).

%% file: data_examples.tex
\vspace{3mm}
\hrule
\begin{example}[\small{Argument and evidence}]\label{ex:data}
\textbf{\\Topic}: We should legalize cannabis
\textbf{\\Argument}: Cannabis can provide relief for a number of ailments without side effects.
\textbf{\\Evidence}: In 1999, a study by the Division of Neuroscience and Behavioral Health found no evidence of a link between cannabis use and the subsequent abuse of other illicit drugs.
\hrule
\end{example}
\vspace{2mm}

%% file: tables/selected_english_data_stats.tex
\begin{table*}[t]
\centering
\renewcommand*{\arraystretch}{0.9}
\begin{tabular}{lcccccccccc}
\toprule
&
\multicolumn{4}{c}{\textbf{\argumentDatasetName}} & 
\multicolumn{6}{c}{\textbf{\evidenceDatasetName}} 
\\
\cmidrule(rl){2-5}
\cmidrule(rl){6-11}
 &
 &
\multicolumn{2}{c}{\emph{\textbf{Stance}}} & 
\emph{\textbf{Quality}} &
\multicolumn{3}{c}{\emph{\textbf{Stance}}} & 
\multicolumn{3}{c}{\emph{\textbf{Evidence}}} 
\\
\cmidrule(rl){3-4}
\cmidrule(rl){5-5}
\cmidrule(rl){6-8}
\cmidrule(rl){9-11}
\bf Set & 
\bf \#T & 
\bf Pro & 
\bf Con & 
\bf \#Args &
\bf \#T & 
\bf Pro & 
\bf Con & 
\bf \#T & 
\bf Ev &
\bf Non-Ev 
\\
\midrule
\bf Train &	49	
& 10,162 & 9,766 & 8,373 & 171	& 5,592	& 3,622 & 174	& 3,522 & 14,275 \\
\bf Dev  &   7	
& 1,564	& 1,497 & 1,329 &  46	& 1,726	& 1,202 & 47 & 1,145 & 3,967 \\ 
\bf Test &	15	
& 3,024	& 2,952 & 2,449 & 100	& 2,614	& 1,209 & 100 & 2,068	& 1,937 \\
\midrule
\bf Total &	71	
& 14,750 & 14,215 & 12,151 & 317	&9,932	& 6,033 & 321	& 6,735	& 20,179 \\
\bottomrule
\end{tabular}
\caption{Statistics of the data selected from the \textbf{\argumentDatasetName} and \textbf{\evidenceDatasetName} datasets and translated into $5$ non-EN languages. 
For the tasks of \emph{\textbf{stance}} classification, argument \emph{\textbf{quality}} prediction and \emph{\textbf{evidence}} detection, 
the table shows: the number of topics (\textbf{\#T}) discussed by the arguments (from \argumentDatasetName) or sentences (from \evidenceDatasetName) for each task; 
the number of \textbf{Pro} and \textbf{Con} arguments or sentences for stance classification; 
the number of arguments (\textbf{\#Args}) for argument quality; 
the number of evidence (\textbf{Ev}) and non-evidence (\textbf{Non-Ev}) sentences for evidence detection. 
} 
\label{tab:selected_english_data_stats}
\end{table*}

%% file: tables/human_generated_corpus.tex
\begin{table*}[t]
\tabcolsep=0.2015cm
\begin{center}
\renewcommand*{\arraystretch}{0.9}
\begin{tabular}{llccccccccccc}

\toprule
& &
\multicolumn{7}{c}{\emph{\textbf{Collection}}} & 
\multicolumn{4}{c}{\emph{\textbf{Evaluation}}} 
\\
\cmidrule(rl){3-9}
\cmidrule(rl){10-13}
 & 
 &
 &
\multicolumn{3}{c}{\emph{\textbf{Stance}}} & 
\multicolumn{3}{c}{\emph{\textbf{Quality}}} &
\multicolumn{3}{c}{\emph{\textbf{Stance}}} & 
\multicolumn{1}{c}{\emph{\textbf{Quality}}} 
\\
\cmidrule(rl){4-6}
\cmidrule(rl){7-9}
\cmidrule(rl){10-12}
\cmidrule(rl){13-13}
\bf \languageColumn & 
 & 
\bf \argumentCountColumn & 
\bf \numLabeledColumn & 
\bf \kappaColumn & 
\bf \supportingArgumentsColumn &
\bf \numLabeledColumn & 
\bf \kappaColumn & 
\bf \highQualityArgumentsColumn &
\bf \#Args & 
\bf Pro & 
\bf Con & 
\bf \#Args
\\
\midrule
\bf Spanish & \bf \spanishShort & 
\esNumUniqueCollectedArguments & 
\statistic{2995} &
\esHumanGeneratedStanceKappaUnfiltered & 
\esHumanGeneratedStancePriorUnfiltered & 
\esLabeledArgumentsFiltered &
\esHumanGeneratedQualityKappaFiltered & 
\esHumanGeneratedQualityPriorFiltered &
\statistic{2541}	& 
\statistic{1337}	& 
\statistic{1204} & 
\statistic{440}
\\
\bf French & \bf \frenchShort & 
\frNumUniqueCollectedArguments & 
\frLabeledArgumentsUnfiltered &
\frHumanGeneratedStanceKappaUnfiltered & 
\frHumanGeneratedStancePriorUnfiltered & 
\frLabeledArgumentsFiltered &
\frHumanGeneratedQualityKappaFiltered & 
\frHumanGeneratedQualityPriorFiltered 
& \statistic{957}	& \statistic{500}	& \statistic{457} & \statistic{556}
\\
\bf Italian & \bf \italianShort & 
\itNumUniqueCollectedArguments & 
\itLabeledArgumentsUnfiltered &
\itHumanGeneratedStanceKappaUnfiltered & 
\itHumanGeneratedStancePriorUnfiltered & 
\itLabeledArgumentsFiltered &
\itHumanGeneratedQualityKappaFiltered & 
\itHumanGeneratedQualityPriorFiltered 
& \statistic{923}	& \statistic{465}	& \statistic{458} & \statistic{586}
\\
\bf German & \bf \germanShort & 
\deNumUniqueCollectedArguments & 
\deLabeledArgumentsUnfiltered &
\deHumanGeneratedStanceKappaUnfiltered & 
\deHumanGeneratedStancePriorUnfiltered & 
\deLabeledArgumentsFiltered &
\deHumanGeneratedQualityKappaFiltered & 
\deHumanGeneratedQualityPriorFiltered 
& \statistic{628}	& \statistic{347}	& \statistic{281} & \statistic{467}
\\
\bf Dutch & \bf \dutchShort & 
\nlNumUniqueCollectedArguments & 
\nlLabeledArgumentsUnfiltered &
\nlHumanGeneratedStanceKappaUnfiltered & 
\nlHumanGeneratedStancePriorUnfiltered & 
\nlLabeledArgumentsFiltered &
\nlHumanGeneratedQualityKappaFiltered & 
\nlHumanGeneratedQualityPriorFiltered
& \statistic{478}	& \statistic{237}	& \statistic{241} & \statistic{264}
\\
\bottomrule
\end{tabular}
\end{center}
\caption{Statistics of the \humanGeneratedArgDatasetName multilingual arguments dataset, collected in five languages (See \sectionRef{par:hg_data}).
On the left are statistics pertaining to its \emph{\textbf{collection}}:
the number of unique arguments collected (\textbf{\argumentCountColumn}); the number of  arguments labeled (\textbf{\numLabeledColumn}) for their \textbf{\textit{stance}} and \textbf{\textit{quality}}; the \agreementName obtained for each task; the average percentage of arguments labeled by each annotator as supporting the topic (\textbf{\supportingArgumentsColumn}) and as high-quality (\textbf{\highQualityArgumentsColumn}).
On the right are statistics describing the \emph{\textbf{evaluation}} data selected from \humanGeneratedArgDatasetName for the stance and quality tasks:
the number of arguments (\textbf{\#Args}) selected for the evaluation of each task; for stance classification, the number of \textbf{Pro} and \textbf{Con} arguments within that selection.
}
\label{tab:human_generated_labeling_stats} 
\end{table*}

%% file: evaluation.tex
\section{Experimental Setup}
\label{empirical_evaluation}

Our experiments are aimed at providing a comparative analysis of the translate-train approach when trained on different subsets of languages, and identifying when that approach is beneficial on the three argumentation tasks.
We begin by describing the setup used in all experiments.

\paragraph{Training Configuration}
We used the BERT-Base multilingual cased model configuration (12-layer, 768-hidden, 12-heads, total of 110M parameters) with a sentence-topic pair input.
Training was performed on one GPU. 

The parameters configuration of the binary classification tasks, namely, stance classification and evidence detection, was:
maximum sequence length of $128$, batch size of $32$, dropout rate of $0.1$ and learning rate of 5e-5. Each model was fine-tuned over $10$ epochs, using a cross-entropy loss function. 
The regression model for argument quality prediction, similar to the one used by \citet{gretz2020quality}, used 
a maximum sequence length of $100$, a batch size of $32$, a dropout rate of $0.1$ and a learning rate of 2e-5. 
Each model was fine-tuned over $3$ epochs, using a mean-squared-error loss function. 
In all cases, the model from the last epoch was selected for evaluation. 

\paragraph{Translate-Train Models}
\label{par:lang_groups}
For each task, mBERT was trained using data translated into one of the target languages (ES, FR, IT, DE and NL).
These per-language models, denoted herein as \emph{TL}, are the simplest application of the translate-train approach.
Two more models were trained for the language families that are represented in the above languages together with English: 
\emph{RM} -- for the Romance languages (ES, FR, IT), and \emph{WG} -- for the West-Germanic languages (EN, DE, NL). 
Each language family model was trained using the data of the languages in that family.
Lastly, a model was trained on data from all 6 languages (denoted \emph{6L}).
To summarize, our evaluation includes $4$ models based on the translate-train approach (\emph{TL}, \emph{6L}, \emph{RM} and \emph{WG}) for each task and TL.

\paragraph{Baselines}
Another mBERT model, denoted \emph{EN}, was trained on the source English data. 
Using this model, the results of two baselines are reported:
(i) zero-shot (denoted \emph{ZS}) -- which passes an input text in a non-English language to the \emph{EN} model, utilizing the cross-lingual transfer capabilities of mBERT; (ii) translate-test (denoted \emph{TT}) -- in which an input text is machine translated into English, and that translation is provided as input to the \emph{EN} model.
The results on English data 
are also reported as a performance benchmark for other languages.
Obviously, for English, the results of the \emph{TL} model and the \emph{ZS} and \emph{TT} baselines are identical.

\input{figures/arguments_stance_results}

\paragraph{Related and Distant Languages}
For each TL, we define two types of models. The \emph{RL} model is the one trained on \emph{related} languages from the same family as the TL, and  \emph{DL} is the model that is trained on languages that are more \emph{distant} from the TL.
In other words, given a TL, the \emph{RL} model refers to the language family that includes the TL, and the \emph{DL} model refers to the other family that does not include the TL.
For example, in the case of the TL being German, \emph{TL} denotes the model trained only on translated data in German; 
\emph{RL} -- the \emph{WG} model trained on the 3 West-Germanic languages (including German); 
\emph{DL} -- the \emph{RM} model trained on the 3 Romance languages.

\paragraph{Evaluation Metrics}
The reported metrics are macro-F1 for the classification tasks (stance classification and evidence detection), and Pearson correlation for the regression setting of argument quality.

%% file: figures/arguments_stance_results.tex
\begin{figure*}
    \centering
    \subfloat[\humanGeneratedArgDatasetName (human-generated)]
    {
    \includegraphics[width=0.46\linewidth, trim={1.2cm, 0.2cm, 1.4cm, 0.2cm}, clip]{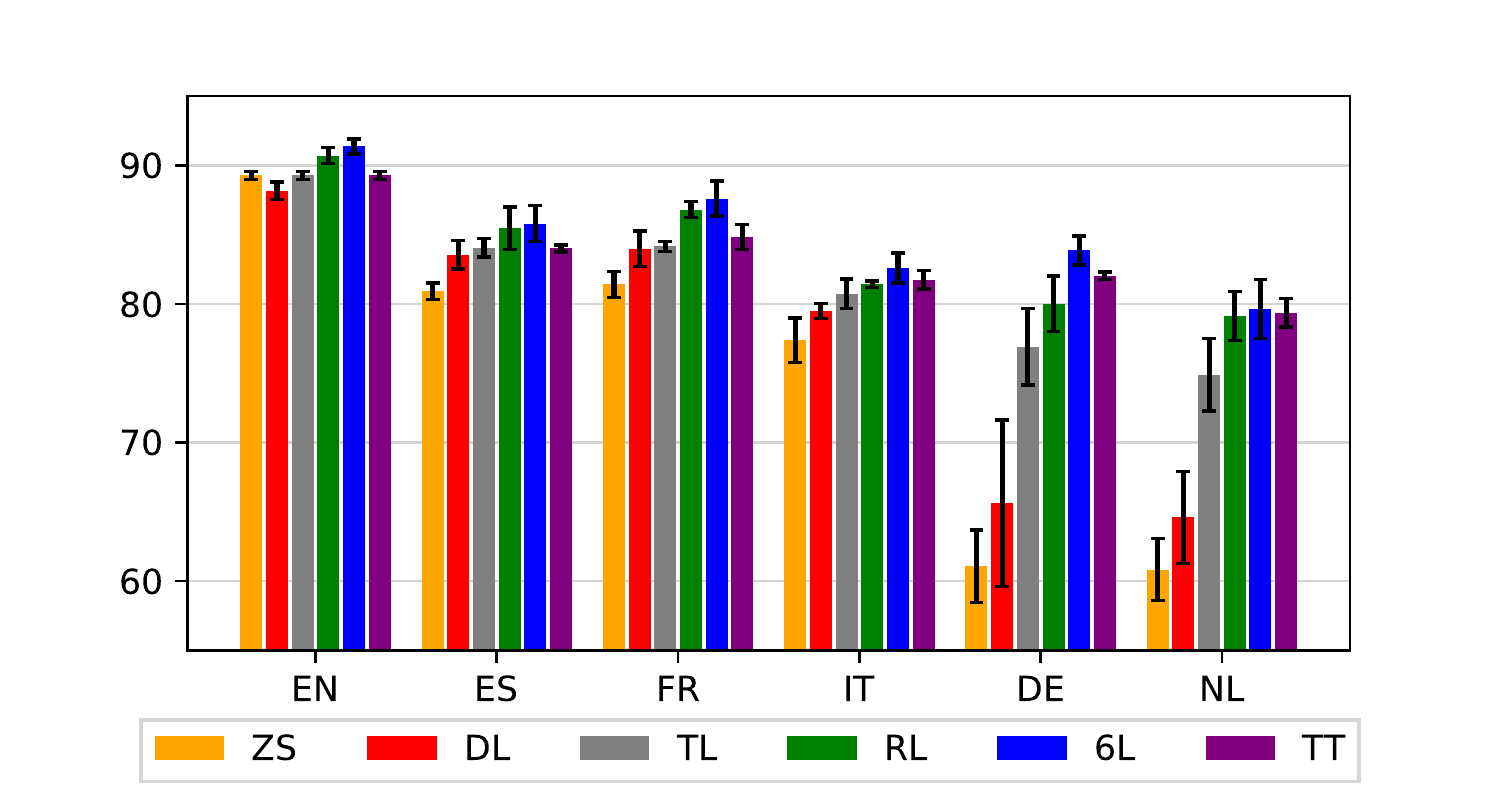}
    \label{fig_sbc_stance_hg}
    }\qquad
    \subfloat[\pseudoTestArgDatasetName (pseudo-test)]
    {
    \includegraphics[width=0.46\linewidth, trim={1.2cm, 0.2cm, 1.4cm, 0.2cm}, clip]{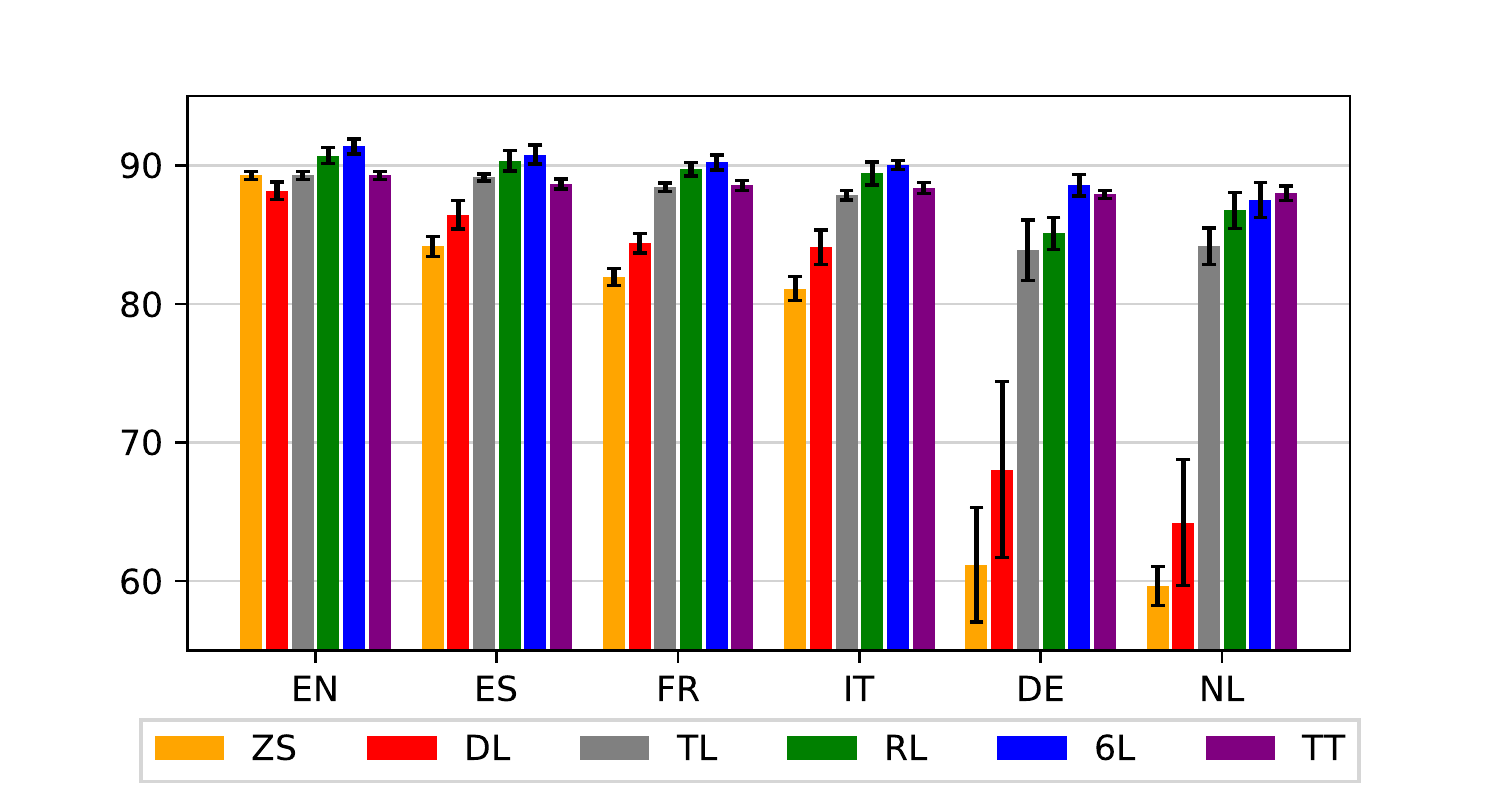}
    \label{fig_sbc_stance_mt}
    }
    \caption{Stance classification
    results on arguments data, showing macro-F1 averaged over 5 evaluation runs, and its standard deviation. 
    The results compare four translate-train methods (\DL, \TL, \RL and \sixL), and two baselines -- zero short (\emph{ZS}) and translate-test (\emph{TT}).
    See \sectionRef{subsec:stance_classification_results} for details.
}
    \label{fig:argument_stance_results}
\end{figure*}

%% file: results.tex
\section{Results}
\label{results}

The results below for arguments (for stance classification on that data and argument quality) are averages over $5$ evaluation runs of randomly initialized models that were trained in the same manner. 
For evidence sentences (stance classification on that data and evidence detection), the results are from a single evaluation run.

\subsection{Stance Classification}
\label{subsec:stance_classification_results}

\paragraph{Arguments} 

\figureRef{fig_sbc_stance_hg} shows the evaluation results on the human-generated arguments of \humanGeneratedArgDatasetName. 
For the non-English languages, the performance over the \emph{ZS} baseline improves when adding translated data, even when that data is from distant languages (\emph{DL}).
The other baseline \emph{TT} is better, yet the best performance is attained by the \emph{6L} 
models
-- significantly so for 3 of the 5 languages (ES, FR and DE).
Notably, ordering the translate-train models by their performance yields the same order for all languages: \emph{DL} is always the worst, followed by \emph{TL}, \emph{RL} and the best performing model \emph{6L}.

Repeating the same experiments on the pseudo-test data of \pseudoTestArgDatasetName resulted in similar trends, depicted in \figureRef{fig_sbc_stance_mt}. 
Further augmentation of the training data with translations to more languages beyond the languages included in the training of the \emph{6L} models (e.g. with 9 or 17 languages) did not significantly improve performance on these languages.
These results are detailed in \tableRef{tab_sbc_stance_mt} within the \supp.

\input{figures/evidence_results}

\paragraph{Evidence} To explore whether the observed trends are data-specific, we repeated the evaluation of the stance classification task with 
the \pseudoTestEviDatasetName dataset of evidence sentences from Wikipedia.
Training was performed on the training set of that corpus (called Wikipedia models).
The results on its pseudo-test set are depicted in \figureRef{fig:evidence_stance_wiki}. 
For the 
non-English
languages, the best performing models are \emph{6L} and \emph{RL}, consistent with our findings for arguments (\figureRef{fig:argument_stance_results}).

The VLD evidence curpus allows further exploration of the stance classification task within the evidence domain.
We trained models on a larger dataset of translated evidence combining the Wikipedia data and selected data from the VLD corpus (called Extended models).
\figureRef{fig:evidence_stance_extended} shows the results obtained using 
these models.
Overall, the performance of the Extended models is significantly better than the performance of the Wikipedia models, in almost all cases, and the \emph{TL} models become competitive even with the \emph{6L} models.

\paragraph{Performance on English}
In comparison with the \emph{ZS} baseline (trained only on English), adding translated training data improves performance on English (leftmost bars in Figures \Ref{fig_sbc_stance_hg} and \ref{fig:evidence_stance_wiki}), for both domains.   
For the evidence data, even translations to distant languages (\emph{DL}) help the Wikipedia model, yet when a lot of training data is available in English (leftmost bars in \ref{fig:evidence_stance_extended}), there is no significant gain from adding translations to the training set.

\paragraph{Summary}
Overall, the best performing models for the stance classification task are the \emph{RL} and \emph{6L} models, in both domains.
Our finding that the \emph{6L} models outperform the \emph{TL} models is consistent with previous results on the XNLI task \citep{hu2020xtreme}.
Interestingly, translated data can be used to improve performance on English as well.

The \emph{ZS} and \emph{TT} baselines are almost always outperformed by the best translate-train model.
However, when a large-scale English corpus is available (\figureRef{fig:evidence_stance_extended}), the \emph{TT} baseline becomes comparable to the best translate-train models.

\subsection{Evidence Detection}
\label{subsec:evidence_detection_results}

To examine whether the above observations are task-specific, we move on to the task of evidence detection.
The results for that task on the 
\pseudoTestEviDatasetName pseudo-test are 
depicted in \figureRef{fig:evidence_detection_wikipedia} (Wikipedia models) and in \figureRef{fig:evidence_detection_extended} (Extended models).

In contrast with the stance results, where in most cases the \emph{6L} models were best, for evidence detection performance may degrade when adding languages. 
The best performing translate-train models are either the \emph{TL} or \emph{RL} models, in all cases.

As in the stance classification results for this corpus, the additional training data used in the Extended models improves performance. 
In addition, the English benchmark results for the Wikipedia models (leftmost bars in  \figureRef{fig:evidence_detection_wikipedia}) can improve by adding languages, or by adding English data (\emph{ZS} bar for EN in \figureRef{fig:evidence_detection_extended}), but there is no significant gain from doing both
(leftmost bars in \figureRef{fig:evidence_detection_extended}).

\subsection{Argument Quality Prediction}
\label{subsec:argument_quality_results}

Moving to our last task, \figureRef{fig_sbc_arg_quality_hg} shows the Pearson correlation results on the human-generated arguments, between the predicted 
quality score and the labeled argument quality score.
In contrast with the stance results, adding data from related languages (the \emph{RL} bars) does not help, and training on the English dataset (the \emph{ZS} bars) is sufficient to obtain a competitive model.\footnote{The performance of the \emph{ZS} model on English is $0.61$ -- comparable to its performance for FR and DE.}
We suspect that the reason for this is that this task is more complex and nuanced than the previous two. 

\section{Analysis}

The performance of a translate-train model may be affected, among other factors, by the translation quality, the extent in which a task-specific label is preserved across that translation, and, for our data, the discussed topic. 
These are analyzed below.

\subsection{Translation Quality Assessment}
We assessed our machine translation quality by computing the BLEU score \cite{papineni2002bleu} between the English arguments from the test set of  \argumentDatasetName
and the same arguments after translation to a TL and back to English. 
For all languages, these scores are above $0.5$ (see \tableRef{tab:mt_bleu}), suggesting the translations are of high quality.

\input{figures/argument_quality}

\subsection{Translated Label Assessment}
\label{subsec:translated_label_assessment}
An important prerequisite for training and evaluating models on automatically translated texts is that the labels of the original texts are preserved under translation, which depends on the specific task at hand.
Example \ref{ex:back-translation} shows one argument and its translation to Spanish and back to English. The translation preserves the original stance,  
but the argument quality is degraded.
Hence, we annotated a sample of the translated texts to assess how often this  happens in each task.
The annotation focuses on one Romance and one West-Germanic language -- Italian and German.

\input{tables/mt_bleu}

\input{quality_example}

\paragraph{Annotation Setup}
$14$ arguments were randomly sampled from each  
topic of the \argumentDatasetName test set, yielding $210$ arguments per language.
Similarly, two sentences were sampled from each  
topic in the \evidenceDatasetName test set, producing $200$ sentences per language. 
All texts were machine translated and human translated by native speakers of each TL.
Both translations of each argument were labeled for their stance and quality, as in \sectionRef{par:hg_data}.
Similarly, the potential evidence sentences were annotated for whether they are valid evidence, and those which are so were also annotated with their stance towards the topic, as in \citet{working-solution}.
In this annotation, TQs were formed from translated texts, with the correct answer taken from the English labels.
The full evidence annotation guidelines are included in the \supp.

\paragraph{Results}
\tableRef{tab:translated_arguments_assessment} shows the assessment results for all tasks and the two languages.
The obtained \agreementName is on par with previously reported values for these tasks (as detailed in \sectionRef{par:hg_data}), though somewhat lower for evidence detection. 
The table further shows Pearson correlation between the original English WA-scores, and the WA-scores of the translated texts.
For evidence detection and argument quality, this computation was performed on texts matching the criteria defined in \sectionRef{subsec:data_selection}. 
The correlation for evidence stance classification was computed on sentences with at least $6$ stance labels on their translated version.

The results show that for both datasets, 
stance is well preserved after translation. 
For evidence detection, the correlation is lower, yet the difference between \machineTranslated and \humanTranslated is small,
suggesting the change in the labels is not due to the automatic translation. 
Thus, the use of translated texts in
these tasks is acceptable, for both training and evaluation.

For argument quality, the correlation is considerably lower, and there is a significant difference between \machineTranslated  and \humanTranslated in \italianShort, as may be expected for such a nuanced task. 
This could be the reason that the translate-train models do not improve performance for this task -- since the quality label is not maintained when an argument is translated, projecting these labels into translated texts introduces significant noise into such training data.

\input{tables/translated_arguments_assessment}

\subsection{Per-Topic Analysis}

\begin{figure}[t]
\includegraphics[width=\linewidth,
trim={1.0cm, 0.2cm, 1.4cm, 0.8cm},
clip]{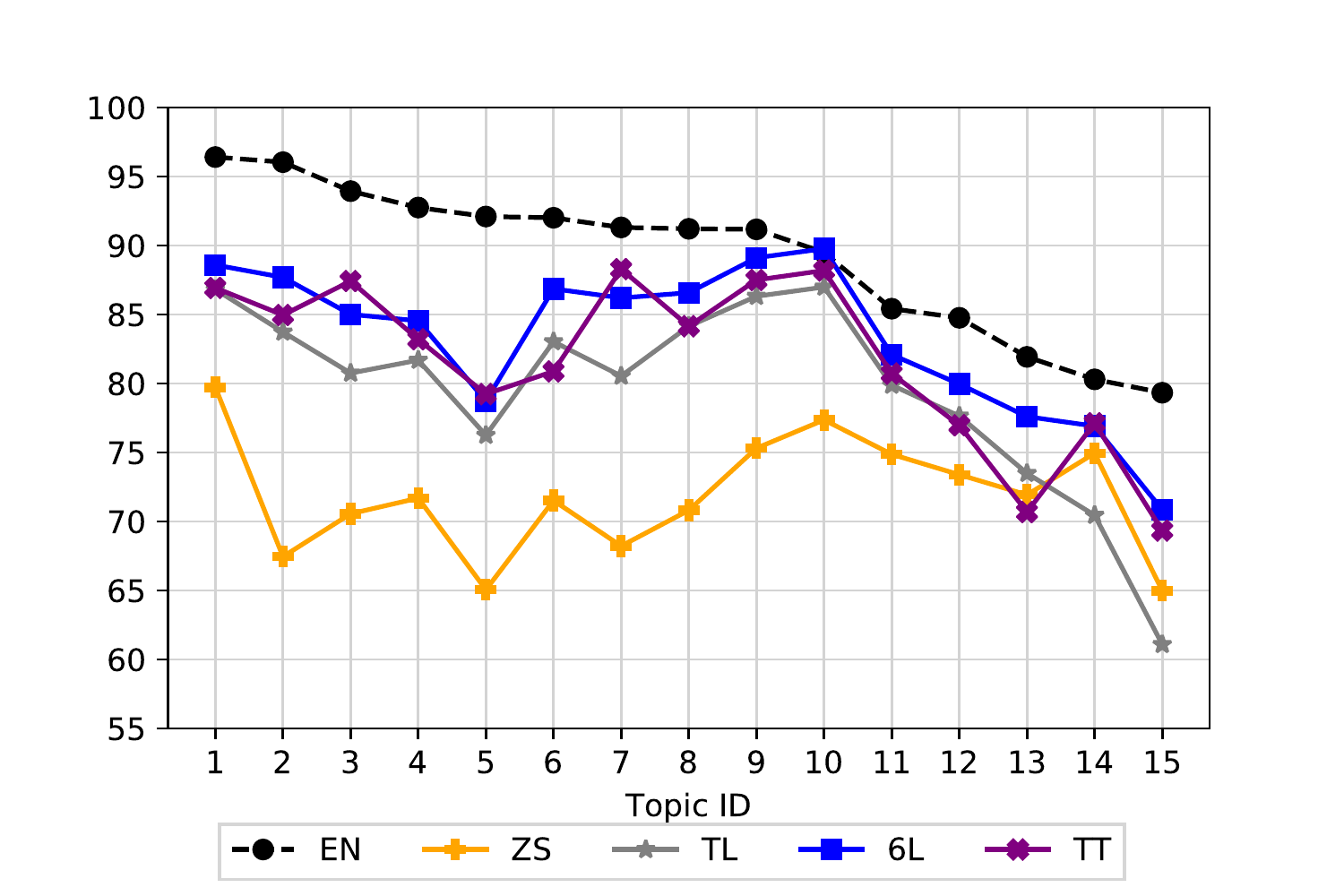}
\caption{Per-topic stance classification average macro-F1 results on 
\humanGeneratedArgDatasetName, averaged over the $5$ non-English languages. 
The topics are ordered by their performance on English with the \emph{EN} model (dashed black line).
}
\label{fig_sbc_stance_hg_avg5tls_per_topic}
\end{figure}

In both of our data domains, arguments and evidence, the texts are relevant to a specific topic, and the obtained performance may depend on that topic in various ways. 
Focusing on stance classification, we measured the per-topic performance on the human-generated arguments of \humanGeneratedArgDatasetName.
\figureRef{fig_sbc_stance_hg_avg5tls_per_topic} shows these results averaged over the $5$ non-English
languages, for the \emph{TL} and \emph{6L} translate-train models, and the \emph{ZS} and \emph{TT} baselines.
The topics are ordered by their performance on English.\footnote{\tableRef{tab:sbc_test_motions} in the \supp lists the topic for each topic ID.}

The results demonstrate the performance variability among the different topics.
For some, the average performance on the 
non-English languages is close to their performance on English (e.g. topics 9 or 10), yet for others it is far from it (e.g. topic 5).  
The performance of the \emph{ZS} baseline is low  
for topics 2 through 8, from which $5$ are discussing imposing a ban. 
This implies that the stance towards the discussed topic, or the  "action" within the topic (e.g. ban, legalize, etc.) may be an important factor.

We further manually analyzed the results on Topic 10, a low-performing
outlier in French for the \emph{ZS} baseline (see \figureRef{fig_sbc_stance_hg_zs_per_topic} in the \supp).
A native speaker examined $3$ batches of $20$ arguments each, containing: 
1) prediction errors from that topic;
2) randomly sampled correct predictions from the same topic;
3) all 4 prediction errors and 16 randomly sampled correct predictions from
the topic with the highest performance (Topic 1).
Within the first batch, $40\%$ of the samples were incoherent or syntactically wrong arguments, compared to only $20\%$ in each of the other two batches.

%% file: figures/evidence_results.tex
\newcommand{\evidenceFigureWidth}[0]{0.46}
\newcommand{\evidenceTopTrim}[0]{0.85cm}
\begin{figure*}
    \centering
    \subfloat[Stance classification, Wikipedia models]
    {
        \includegraphics[width=\evidenceFigureWidth\linewidth, trim={1.2cm, 0cm, 1.4cm, \evidenceTopTrim}, clip]{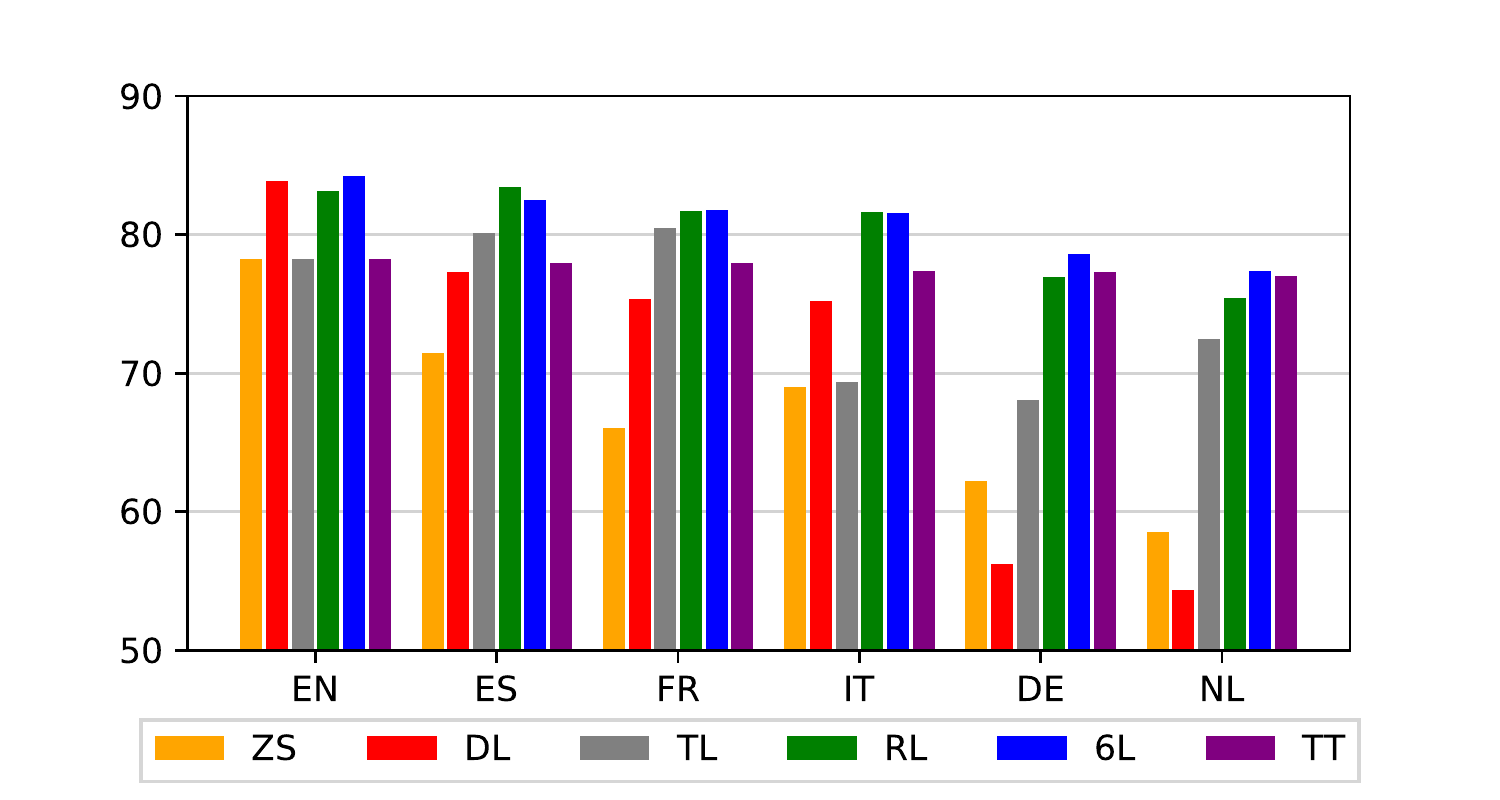}
        \label{fig:evidence_stance_wiki}
    }
    \qquad
    \subfloat[Stance classification, Extended models]
    {
        \includegraphics[width=\evidenceFigureWidth\linewidth, trim={1.2cm, 0cm, 1.4cm, \evidenceTopTrim}, clip]{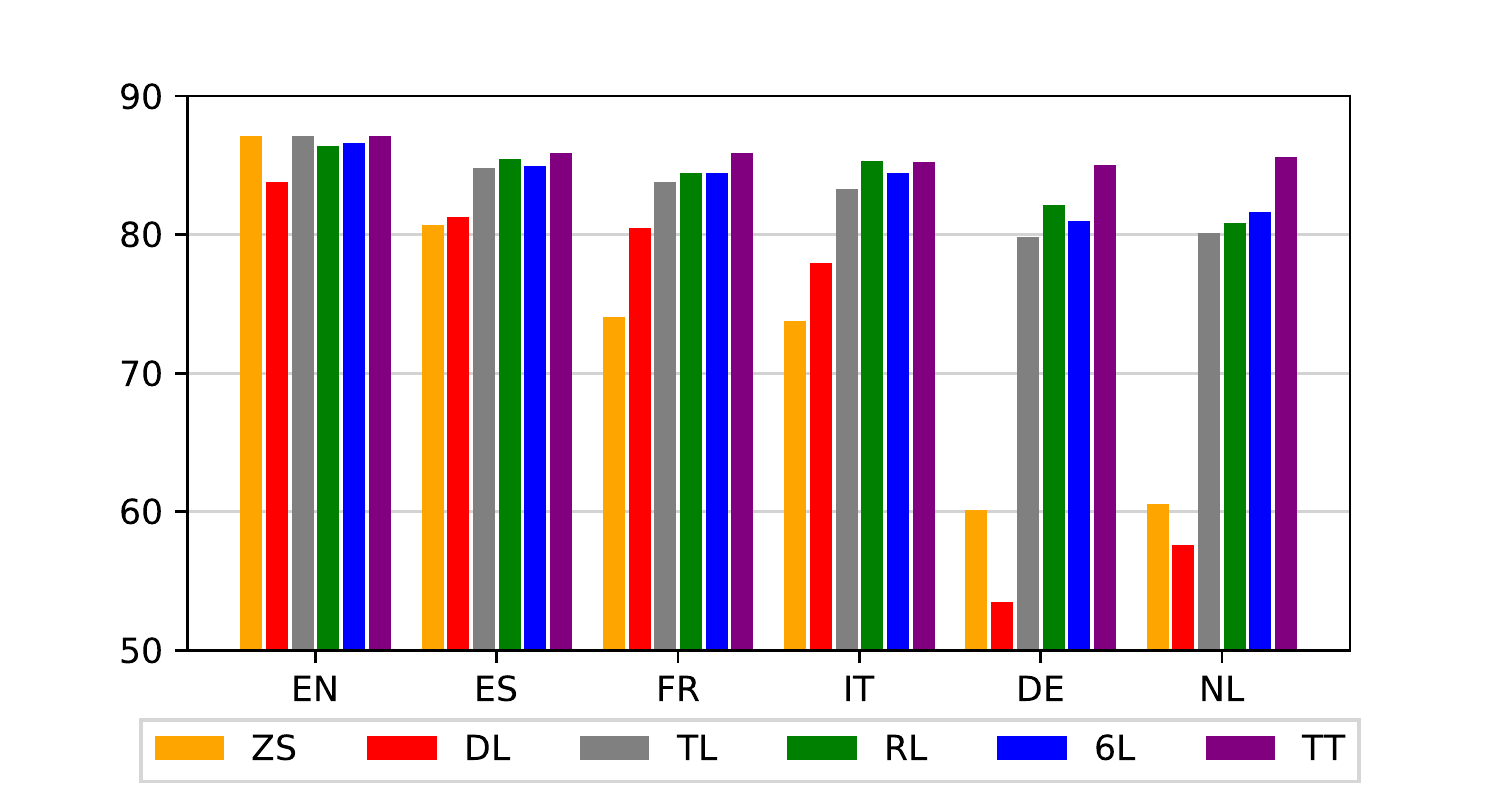}
        \label{fig:evidence_stance_extended}
    }
    
    \subfloat[Evidence detection, Wikipedia models]
    {
        \includegraphics[width=\evidenceFigureWidth\linewidth, trim={1.2cm, 0cm, 1.4cm, \evidenceTopTrim}, clip]{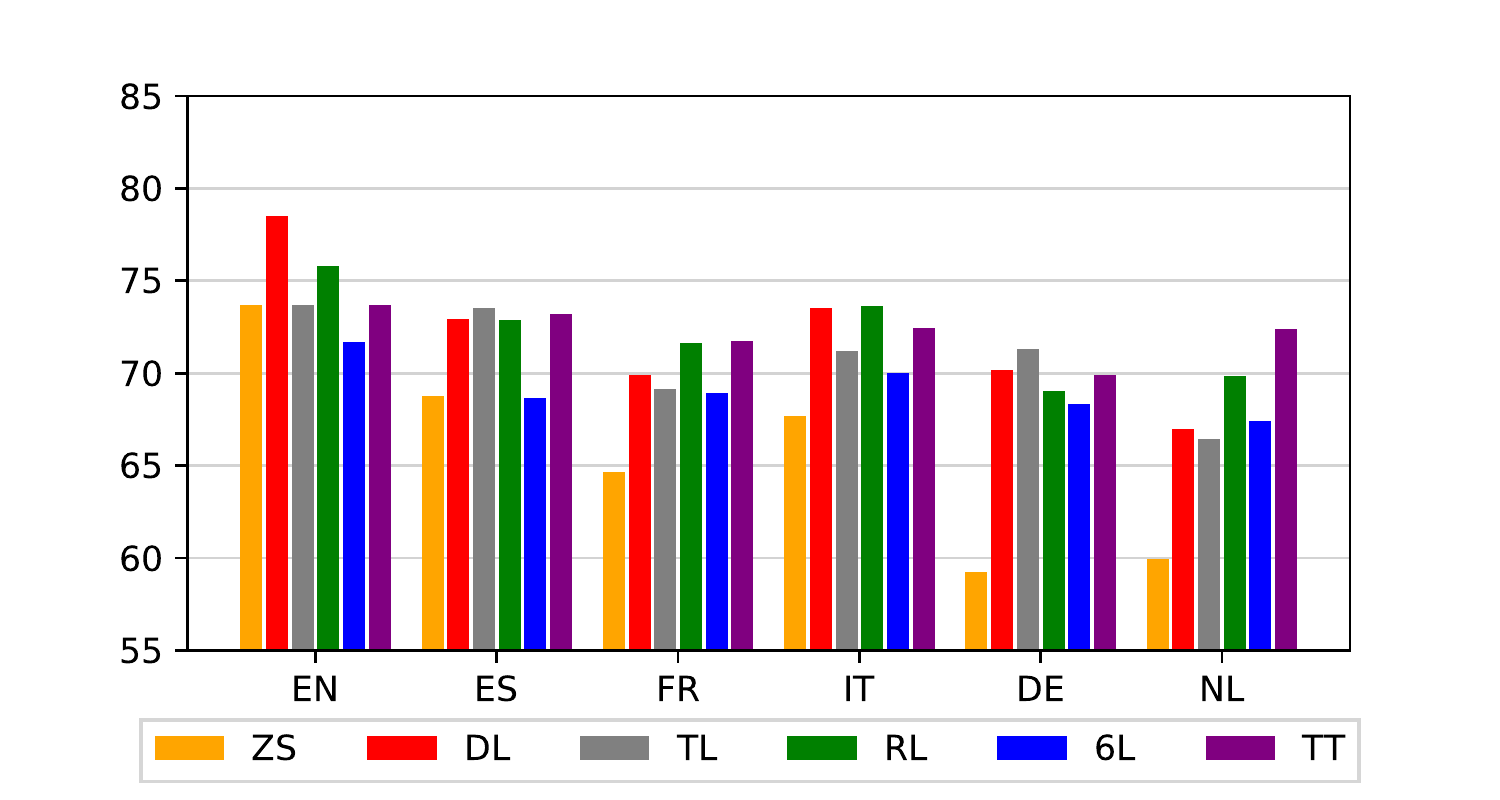}
        \label{fig:evidence_detection_wikipedia}
    }
    \qquad
    \subfloat[Evidence detection, Extended models]
    {
        \includegraphics[width=\evidenceFigureWidth\linewidth, trim={1.2cm, 0cm, 1.4cm, \evidenceTopTrim}, clip]{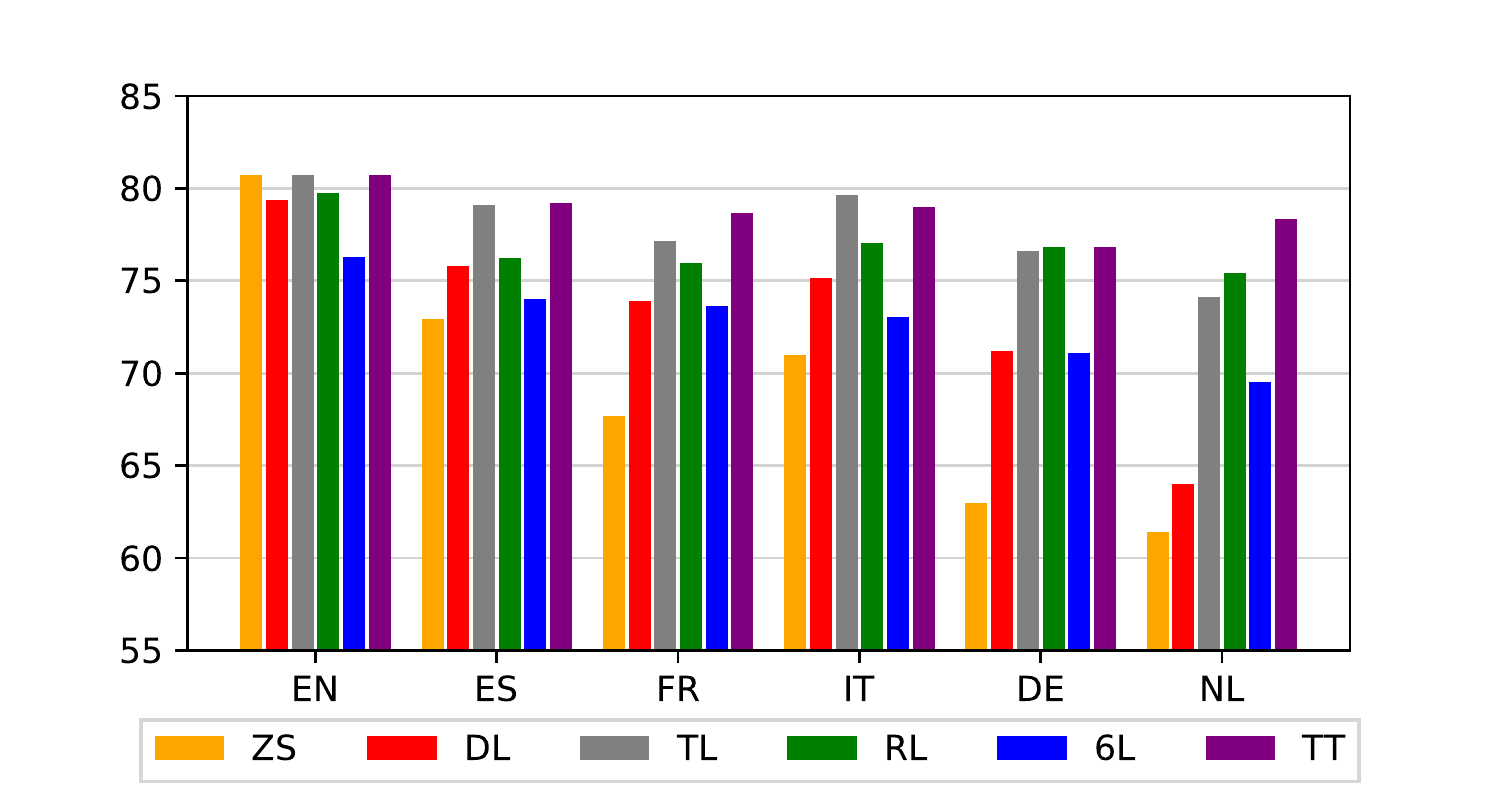}
        \label{fig:evidence_detection_extended}
    }
    \caption{Stance classification (top) and evidence detection (bottom) macro-F1 results on the \pseudoTestEviDatasetName pseudo-test set
    with Wikipedia models (left) and Extended models (right).
    The results compare four translate-train methods (\DL, \TL, \RL and \sixL), and two baselines -- zero short (\emph{ZS}) and translate-test (\emph{TT}). 
    See \sectionRef{subsec:evidence_detection_results} for details.
    }
     \label{fig:evidence_results}
\end{figure*}

%% file: figures/argument_quality.tex
\begin{figure}
\squeezeup
\includegraphics[width=\linewidth, trim={0.8cm, 0.2cm, 1.4cm, 0.2cm}, clip]{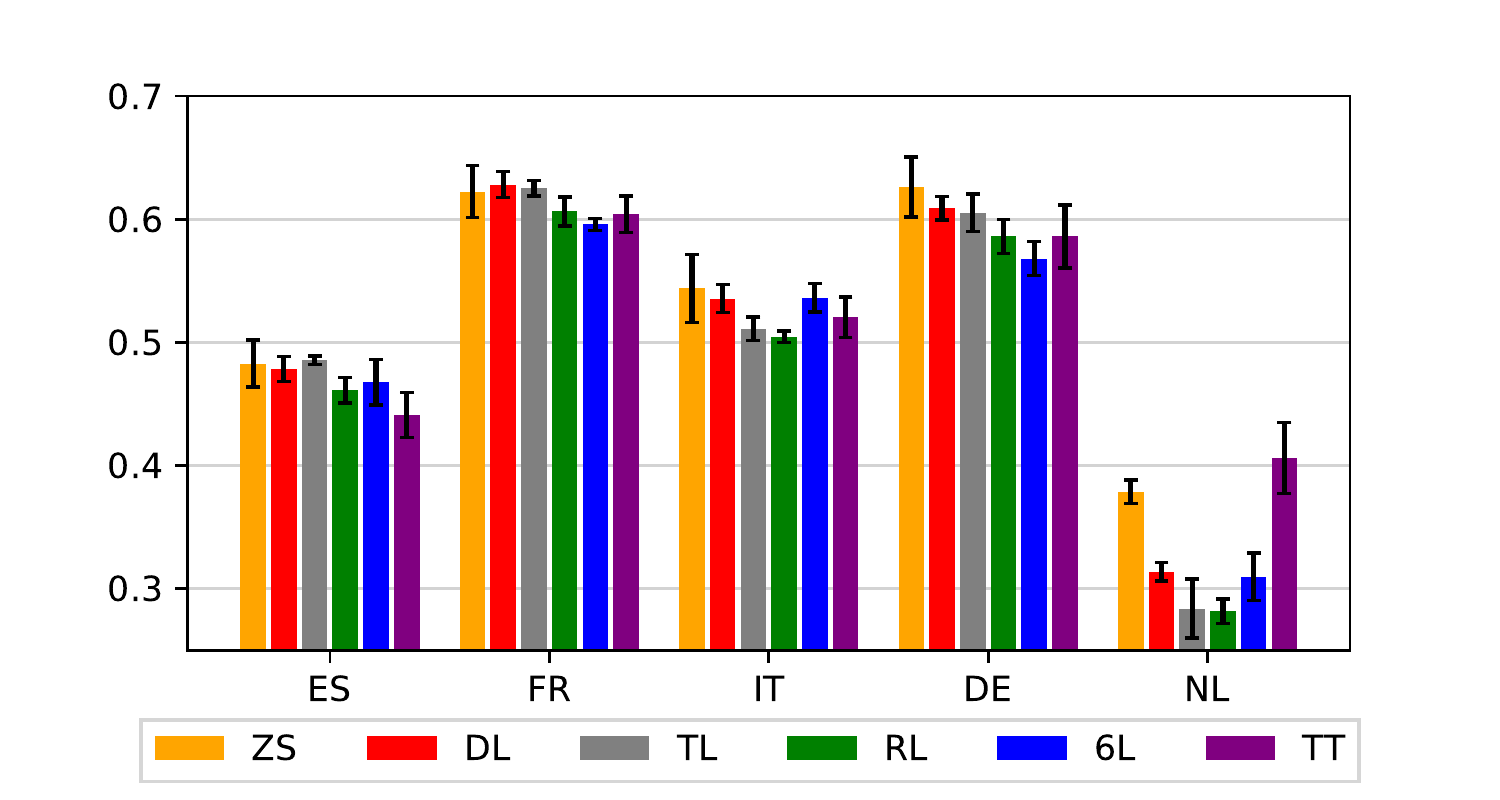}
\caption{
    Argument quality prediction results (Pearson correlation) on the \humanGeneratedArgDatasetName dataset (See \sectionRef{subsec:argument_quality_results}).
}
\label{fig_sbc_arg_quality_hg}
\end{figure}

%% file: tables/mt_bleu.tex
\begin{table}
\centering
\begin{tabular}{ccccc}
\hline
\bf{ES} & \bf{FR} & \bf{IT} & \bf{DE} & \bf{NL} \\
\hline
0.59 &	0.51  &	0.63 &	0.52 &	0.55 \\
\hline
\end{tabular}
\caption{Per-language BLEU scores between the 
\argumentDatasetName
test set arguments and their back-translation to English.}
\label{tab:mt_bleu}
\end{table}

%% file: quality_example.tex
\vspace{3mm}
\hrule
\begin{example}[Translation quality]\label{ex:back-translation}
\textbf{\\Topic}: We should ban algorithmic trading
\textbf{\\English argument}: Algorithmic trading results in unfair advantages for those able to access it to the detriment of ordinary investors.
\textbf{\\Back-translation}: The algorithmic trading of results in unjust advantages for those able to access it to the detriment of common investors.
\hrule
\end{example}
\vspace{1mm}

%% file: tables/translated_arguments_assessment.tex
\begin{table}[t]
\tabcolsep=0.119cm
\begin{center}
\renewcommand*{\arraystretch}{0.93}
\begin{tabular}{l c c c c c c}

\toprule
 & 
\multicolumn{2}{c}{\emph{\textbf{\kappaColumn}}} &
 \humanTranslated &
 \machineTranslated &
 \humanTranslated &
 \machineTranslated 
\\
\cmidrule(rl){2-3}
\cmidrule(rl){4-4}
\cmidrule(rl){5-5}
\cmidrule(rl){6-6}
\cmidrule(rl){7-7}
\bf \taskColumn & 
\bf \germanShort &
\bf \italianShort &
\bf \germanShort &
\bf \germanShort &
\bf \italianShort &
\bf \italianShort 
\\
\midrule
\bf \argumentStanceCorrelationColumn & 
\deTranslatedStanceKappaFiltered &
\itTranslatedStanceKappaFiltered &
\deHTStancePearsonCorrelation &
\deMTStancePearsonCorrelation &
\itHTStancePearsonCorrelation &
\itMTStancePearsonCorrelation 
\\
\bf \evidenceStanceCorrelationColumn & 
\deEvidenceStanceKappa &
\itEvidenceStanceKappa &
\deHTEvidenceStanceSixLabelsPearsonCorrelation &
\deMTEvidenceStanceSixLabelsPearsonCorrelation &
\itHTEvidenceStanceSixLabelsPearsonCorrelation &
\itMTEvidenceStanceSixLabelsPearsonCorrelation 
\\
\bf \qualityCorrelationColumn &
\deTranslatedQualityKappaFiltered &
\itTranslatedQualityKappaFiltered &
\deHTQualityStrongAllPearsonCorrelation &
\deMTQualityStrongAllPearsonCorrelation &
\itHTQualityStrongAllPearsonCorrelation &
\itMTQualityStrongAllPearsonCorrelation 
\\
\bf \evidenceCorrelationColumn &
\deEvidenceDetectionKappa &
\itEvidenceDetectionKappa &
\deMTStrongEvidenceDetectionPearsonCorrelation &
\deHTStrongEvidenceDetectionPearsonCorrelation &
\itHTStrongEvidenceDetectionPearsonCorrelation &
\itMTStrongEvidenceDetectionPearsonCorrelation 
\\
\bottomrule
\end{tabular}
\end{center}
\caption{
Translated labels assessment results for two languages and all tasks: 
stance classification on arguments (\textbf{\argumentStanceCorrelationColumn}) or evidence (\textbf{\evidenceStanceCorrelationColumn}),
argument quality (\textbf{\qualityCorrelationColumn}) and evidence detection (\textbf{\evidenceCorrelationColumn}).
The results show the \agreementName obtained in the annotations, and Pearson correlations between the original English labels and the labels of human (\humanTranslated) and machine (\machineTranslated) translated texts.
}
\label{tab:translated_arguments_assessment} 
\vspace{-2mm}
\end{table}

%% file: conclusions.tex
\section{Conclusions}
\label{sec:conclusions}

We have examined the translate-train paradigm for three multilingual argument mining tasks: stance classification, evidence detection, and argument quality, evaluating a wide range of multilingual models on machine-translated and human-authored data. 
These tasks differ in their complexity, as reflected in the agreement of annotators on the correct label, the extent to which this label is preserved across translation, and, ultimately, in the accuracy of the models. 

Accordingly, our results show that the translate-train approach is well suited for stance classification, as performance improves when 
augmenting the English training data with automatic translations from other languages.
For evidence detection, adding data from the target language or related languages improves performance, yet adding more languages is not helpful.

For both tasks on the evidence data, adding more English 
training data improves performance. 
In these cases, augmenting the large English training set with data of other languages only leads to a marginal gain
for stance classification, and even degrades performance for evidence detection.

In contrast with the above two tasks, the results on argument quality show that training only on English is at least as good, if not better, than any translate-train model.
This is reflected by the clearly opposite trends observed in \figureRef{fig_sbc_arg_quality_hg} vs. those observed in \figureRef{fig_sbc_stance_hg}.

Taken together, our results confirm the validity of the common translate-train paradigm for argument mining tasks such as stance and evidence detection, for which the label is relatively well preserved under translation. However, for the more subtle argument-quality task, where the label -- as might be expected -- is far less preserved, a new approach might be needed.
Future work might wish to explore how translation can preserve not only the semantics of texts, but also finer aspects that contribute to its quality.

%% file: Appendix/appendix.tex
\appendix

\section{Appendices}

\subsection{Additional Results}
\label{sec:appendix}

\subsubsection{Stance Classification on Machine-Translated Arguments}
\label{appendix:stance_mt_17l_results}

As described in \sectionRef{empirical_evaluation},
our evaluations were conducted on 6 European languages (EN, ES, FR, IT, DE, and NL). 
Models were trained for several language groups: \emph{RM} -- for the Romance languages, \emph{WG} -- the West-Germanic, and \emph{6L} -- a model that covers all the TLs in our evaluation.

We further explored the translate-train approach by augmenting the training data of our models with machine-translated data of other language families. 
First, we trained a model for the North-Germanic family (\emph{NG}) with three languages -- Danish (DA), Swedish (SV), and Norwegian (NB).
Next, we combined the Romance languages with  the two German families (\emph{RM, WG, and NG}), and created the \emph{9L} model with $9$ languages.
Finally, we trained a model with a relatively large number of languages ($17$) and a variety of language families. This model, denoted \emph{17L}, consists of all the languages in \emph{9L} and $8$ additional languages: Slavic languages -- Polish (PL), Slovak (SK), Russian (RU); Semitic -- Arabic (AR), Hebrew (HE); and Chinese/Japonic -- Simplified Chinese (ZH), Traditional Chinese (ZT), and Japanese (JA). 

The stance classification results on 
the \pseudoTestEviDatasetName pseudo-test for
all $17$ languages using all the aforementioned models are presented in \tableRef{tab_sbc_stance_mt}. 
We see that expanding the training set beyond the six languages in \emph{6L} by adding more distant languages, as in the \emph{9L} and \emph{17L} models, does not significantly improve the performance on English. 
On average, 
training on the TL is better than training on the original English arguments (average performance over all $17$ languages 
is $73.7\%$ with \emph{EN} and $86.5\%$ with \emph{TL}). Training on all $17$ languages tends to yield the best performance (with an average of $88.9\%$), though training on a subset of them is often nearly as good - and sometimes even better, especially on the \emph{6L} and \emph{9L} groups.

\input{tables/sbc_stance_mt}

\subsubsection{Per-Topic Analysis Information}
\tableRef{tab:sbc_test_motions} contains a list of $15$ topics that are included the test set of the \argumentDatasetName dataset, along with their IDs used during error analysis.

\input{tables/sbc_test_motions_ids}

\begin{figure}[th]
\includegraphics[width=\linewidth,
trim={1.0cm, 0.2cm, 1.4cm, 0.8cm},
clip]{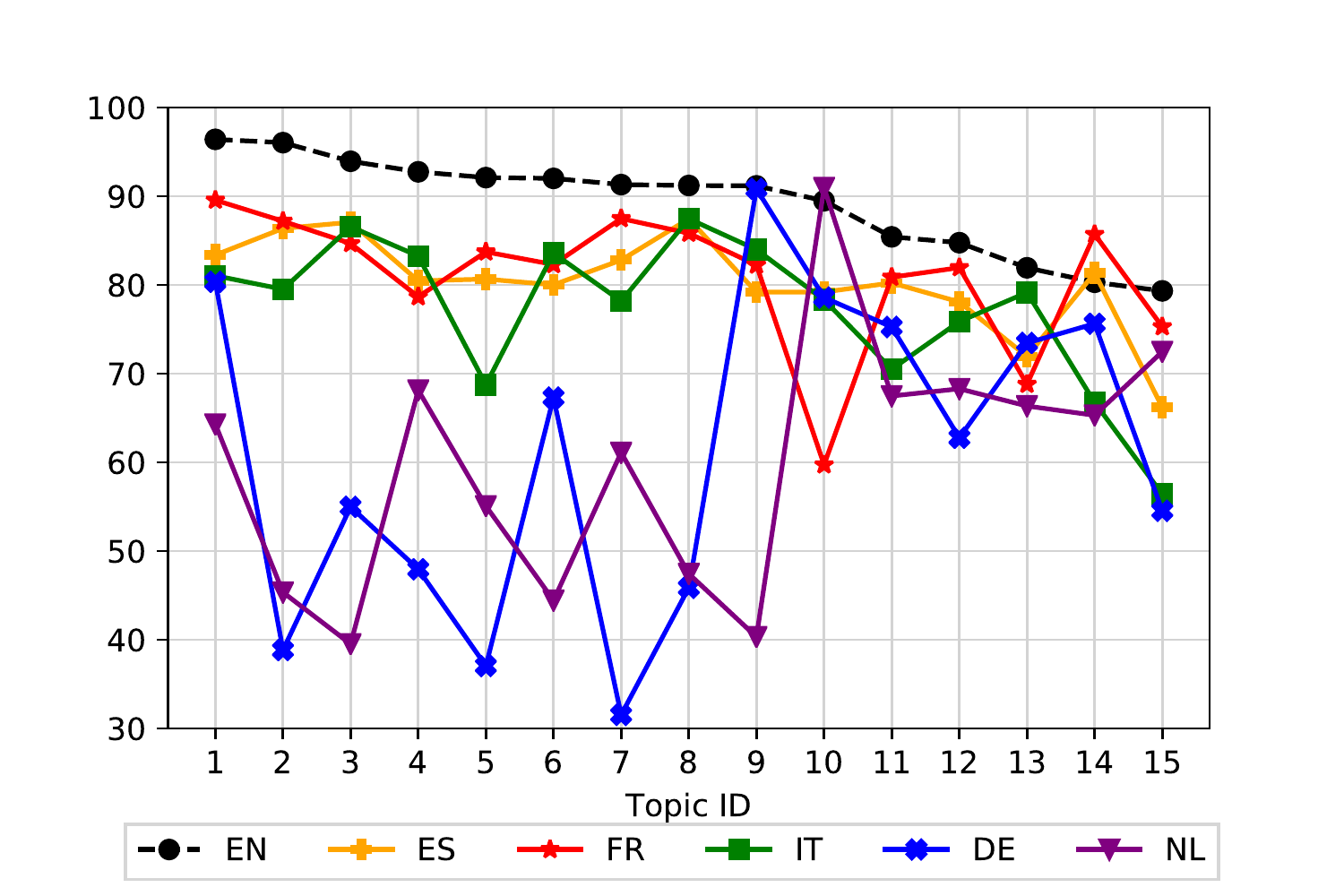}
\caption{Per-topic stance classification
average macro-F1 results per language on \humanGeneratedArgDatasetName test set for the \emph{ZS} baseline. The dashed black line is the performance on English.}
\label{fig_sbc_stance_hg_zs_per_topic}
\end{figure}

\figureRef{fig_sbc_stance_hg_zs_per_topic} shows 
per-language results of the zero-shot baseline on the $15$ topics of the \humanGeneratedArgDatasetName evaluation dataset.
As indicated by the overall results on the same data in \figureRef{fig_sbc_stance_hg}, we see high performance on the three Romance languages consistently across most topics, and low performance on DE and NL for about half of the topics.   

\subsection{Annotation Details}
\label{appendix:annotation_details}
This section describes further annotation details, such as the adjustment of the argument assessment annotation task to multiple languages, and the guidelines used in each annotation task.
\input{Appendix/annotation_scaling}
\input{Appendix/argument_authoring_guidelines}
\input{Appendix/argument_assessment_guidelines}
\input{Appendix/evidence_assessment_guidelines}

%% file: tables/sbc_stance_mt.tex
\begin{table*}
\small
\centering
\addtolength{\tabcolsep}{-2.5pt}
\begin{tabular}{lccccccccccccccccc}
\toprule
\bf{Model} & \bf{EN} & \bf{DE} & \bf{NL} & \bf{ES} & \bf{FR} & \bf{IT} & \bf{DA} & \bf{SV} & \bf{NB} & \bf{PL} & \bf{SK} & \bf{RU} & \bf{AR} & \bf{HE} & \bf{ZH} & \bf{ZT} & \bf{JA} \\
\cmidrule(rl){1-1}
\cmidrule(rl){2-4}
\cmidrule(rl){5-7}
\cmidrule(rl){8-10}
\cmidrule(rl){11-13}
\cmidrule(rl){14-15}
\cmidrule(rl){16-18}
\emph{\textbf{EN}} & 
\textit{89.3} & \textit{61.2} & \textit{59.7} &	\textit{84.2} &	\textit{82.0} & \textit{81.1} &
69.0 & 72.8 & 66.7 & 74.8 & 74.3 & 73.6 & 62.7 & 71.9 & 81.4 & 80.4 & 67.7 \\
\emph{\textbf{TL}} & 
\textit{89.3} & \textit{83.9} & \textit{84.2} & \textit{89.1} & \textit{88.4} & \textit{87.9} &
84.8 & 86.6 & 87.0 & 84.8 & 87.5 & 86.8 & 85.3 & 85.5 & 87.7 & 87.5 & 83.6 \\
\emph{\textbf{ES, FR, IT}} & 
\textit{88.2}	& \textit{68.0} &	\textit{64.2} & \textit{90.3} &	\textit{89.7} & \textit{89.4} &
65.2 & 74.3 & 74.4 & 73.7 & 72.8 & 75.0 & 71.1 & 77.0 & 83.1 & 84.0 & 70.5 \\
\emph{\textbf{EN, DE, NL}} & 
\textit{90.7} & \textit{85.1} & \textit{86.8} & \textit{86.4} &	\textit{84.4}	 & \textit{84.1} &
79.8 & 76.5 & 79.6 & 78.1 & 75.9 & 79.1 & 68.7 & 67.4 & 80.8 & 81.4 & 70.9 \\
\emph{\textbf{SV, NB, DA}} & 
\textit{83.7} & \textit{64.0} & \textit{68.3} & \textit{82.8} & \textit{80.1} & \textit{79.0} &
88.8 & 88.9 & 88.6 & 72.8 & 73.8 & 76.7 & 64.1 & 71.1 & 79.1 & 79.6 & 67.3 \\
\emph{\textbf{PL, SK, RU}} & 80.5 & 60.1 & 56.5 & 80.8 & 82.7 & 81.5 & 75.8 & 77.4 & 74.2 & 86.8 & 88.3 & 88.4 & 70.9 & 75.4 & 82.1 & 81.5 & 66.3 \\
\textbf{AR, HE} & 81.2 & 60.7 & 54.7 & 78.1 & 79.2 & 78.9 & 66.1 & 71.9 & 63.0 & 75.7 & 73.6 & 74.2 & 85.7 & 86.2 & 80.0 & 81.3 & 66.5 \\
\emph{\textbf{ZH, ZT, JA}} & 81.3 & 63.8 & 55.6 & 81.3 & 69.3 & 78.7 & 61.6 & 65.6 & 61.9 & 72.4 & 70.9 & 75.8 & 62.9 & 71.5 & 88.6 & 88.3 & 85.8 \\
\emph{\textbf{6L}} & 
\textit{91.4} & \textit{\textbf{88.6}}	& \textit{87.5} & 90.8 & \textit{90.2} & \textit{\textbf{90.0}} &
70.6 & 81.2 & 75.1 & 79.4 & 73.2 & 83.4 & 69.2 & 70.2 & 81.9 & 82.3 & 73.2 \\
\emph{\textbf{9L}} & 
\textit{91.3} & \textit{86.6} & \textit{\textbf{88.8}} & \textit{\textbf{90.9}} & \textit{90.3} & \textit{\textbf{90.0}} &
\textbf{89.7} & \textbf{89.0} & 89.7 & 76.7 & 72.3 & 81.7 & 70.6 & 71.7 & 83.0 & 84.1 & 73.2 \\
\emph{\textbf{17L}} & 
\textit{\textbf{91.5}} & \textit{\textit{86.8}} & \textit{\textbf{88.8}} & \textit{90.7} & \textit{\textbf{90.5}} & \textit{\textbf{90.0}} &
89.3 & 88.9 & \textbf{90.0} & \textbf{88.3} & \textbf{89.1} & \textbf{88.9} & \textbf{87.8} & \textbf{86.9} & \textbf{88.9} & \textbf{88.8} & \textbf{86.8} \\
\bottomrule
\end{tabular}
\addtolength{\tabcolsep}{2.5pt}
\caption{Stance classification macro-F1 results 
on the \argumentDatasetName test set 
in English (leftmost column) and on the 
\pseudoTestArgDatasetName 
evaluation
set in 16 
languages with models trained on various language groups (see 
\sectionRef{appendix:stance_mt_17l_results}). 
The results in italics are showing
averages of 5 training runs. 
The other results are from a single training run of each model.}
\label{tab_sbc_stance_mt}
\vspace{-2mm}
\end{table*}

%% file: tables/sbc_test_motions_ids.tex
\begin{table}[ht]
\centering
\begin{tabular}{cp{0.8\linewidth}}
\hline
\bf{ID} & \bf{Topic} \\
\hline
1	& We should abolish the Olympic Games \\
2	& We should ban factory farming \\
3	& We should ban algorithmic trading \\
4	& We should ban targeted killing \\
5	& We should prohibit school prayer \\
6	& We should ban private military companies \\
7	& We should adopt libertarianism \\
8	& We should ban missionary work \\
9	& Social media brings more harm than good \\
10	& We should legalize cannabis \\
11	& We should abolish the three-strikes laws \\
12	& We should prohibit women in combat \\
13	& Holocaust denial should be a criminal offence \\
14	& The use of public defenders should be mandatory \\
15	& We should adopt atheism \\
\hline
\end{tabular}
\caption{Topics and IDs of the \argumentDatasetName test set.}
\label{tab:sbc_test_motions}
\end{table}

%% file: Appendix/annotation_scaling.tex
\subsubsection{Multilingual Argument Assessment} 
While \citet{gretz2020quality} mention 
using a group of annotators with whom they have worked before for the assessment of arguments written in English, no such group was available to us for the non-English languages.
In addition, no test questions (TQs) were available, since they are typically formed from existing labeled data.
Initially, the first issue was addressed by relying on workers from appropriate countries, and the second by using machine-translated arguments from \argumentDatasetName, with a high-confidence label \emph{in English}, as TQs. 
At first, since the quality label is sensitive to translation (as described in \sectionRef{subsec:translated_label_assessment}) such TQs were limited to stance. 

A pilot on Spanish arguments showed a good \agreementName for stance ($0.71$), yet a low value for quality ($0.04$). 
The results showed that many of the annotators labeled a vast majority ($>$\percentage{80}) of the arguments as high-quality, 
even though they were instructed to consider only half as such.
Therefore, only those labeling $\leq80\%$ of arguments as high-quality were allowed further work. Others were excluded and their argument quality answers were ignored.

A second pilot extended this procedure to other languages.  However, the size of the workforce meeting the above criteria was small for \germanShort, \frenchShort and \dutchShort, preventing progress altogether for the last two.
This required integrating TQs for quality question despite the risk of the quality label changing due to the automatic translation.
To mitigate that risk, one of the authors carefully monitored each annotation task, reviewed TQs which many annotators answered incorrectly, and disabled those in which the translation introduced errors in the correct label or made the text unclear.

%% file: Appendix/argument_authoring_guidelines.tex
\subsubsection{Argument Authoring Guidelines}
Below is an example of the argument authoring guidelines for German. The guidelines for the other languages were similar.
\\ \\ 
PLEASE READ: \\ \\
All your submitted arguments will be assessed for their quality. For each argument determined as a high-quality one, you will receive a bonus of up to 0.4\$.

\subsection*{Overview}
In the following task you are presented with a debatable topic, to which you should suggest high quality supporting/contesting arguments in \underline{German}.
\\ \\
A supporting/contesting argument will be considered as a high-quality one, if a person preparing a speech to support/contest the topic, respectively, will be likely to use this argument as is in her speech.
\\ \\
\textbf{Note:} Copying texts from the web or elsewhere is prohibited. The content you provide must be written by you in your own language. 
\\ \\
\textbf{Requirements}
\begin{itemize}
\vspace{-0.2cm}\item The argument must be phrased in \underline{German}.
\vspace{-0.2cm}\item The argument must either clearly support or clearly contest the topic.
\vspace{-0.2cm}\item You should write a single argument in each text box.
\end{itemize}

%% file: Appendix/argument_assessment_guidelines.tex
\subsubsection{Argument Assessment Guidelines}
In this annotation, the guidelines for all languages were the same.
\\ \\
In the following task you should answer two questions concerning an argument suggested in the context of a debatable topic.

\begin{enumerate}
    \vspace{-0.2cm}\item What is the stance of the argument towards the topic? (supporting, contesting or neutral)
    \vspace{-0.2cm}\item For someone with this stance towards the topic, is this a high-quality argument to use? (yes or no)
\end{enumerate}
IMPORTANT! For the second question please answer "YES" only for high-quality arguments, and only for about half of the time.
\\ \\ 
Your answers will be monitored not only using test questions. If you are interested in participating in future similar tasks, please answer thoroughly.

%% file: Appendix/evidence_assessment_guidelines.tex
\subsubsection{Evidence Assessment Guidelines}
Below is an example of the evidence assessment guidelines for German.
The guidelines for the other languages were similar.

\section*{General instructions}

In this task you are given a topic and evidence candidates for the topic. The candidates are in German. Consider each candidate independently. For each candidate please select Accept if and only if it satisfies ALL the following criteria:

\begin{itemize}
    \vspace{-0.2cm}\item The candidate clearly supports or clearly contests the given topic. A candidate that is neutral towards the topic should not be accepted.
    \vspace{-0.6cm}\item The candidate represents a coherent, stand-alone statement, that one can articulate (nearly) “as is” while discussing the topic, with no need to change/remove/add more than two words.
    \vspace{-0.2cm}\item The candidate represents valuable evidence to convince one to support or contest the topic. Namely, it is not merely a belief or merely a claim, rather it provides an indication whether a belief or a claim is true. A candidate which presents detailed information (typically quantitative) that clearly support or clearly contest the topic, should be accepted.
\end{itemize}
If you select Accept, you should further indicate whether the evidence supports the topic (Pro) or contests it (Con).
\\
\textbf{Note:} if you are unfamiliar with the topic, please briefly read about it in a relevant data source like Wikipedia.

\section*{Examples}

The following examples outline several candidates along with their suggested annotations; please read all these examples before performing the task.
\\ \\
\textbf{Topic:} We should ban the sale of violent video games to minors.
\\ \\
\underline{Example 1}
\\\vspace{-2mm}\\   
\textit{The research clearly suggests that, among other risk factors, exposure to violent video games can lead to aggression and other potentially harmful effects.}
\\\vspace{-3mm}\\ 
\textbf{Annotation:} Accept – Pro
\\\vspace{-3mm}\\ 
\textbf{Note:} even though the text is not explicitly referring to the proposed ‘ban’ policy, it should still be accepted, since highlighting the negative aspects of violent video games can be used to support the suggested ban.
\\ \\
\underline{Example 2}
\\\vspace{-2mm}\\ 
\textit{A university of Oxford study negates the idea that violent video game content leads to violence.}
\\\vspace{-3mm}\\ 
\textbf{Annotation:} Accept - Con
\\\vspace{-3mm}\\ 
\textbf{Note:} here as well, even though the proposed ‘ban’ policy is not explicitly mentioned, the text should be accepted since clearly it can be used to contest the suggested ban.
\\ \\
\underline{Example 3}
\\\vspace{-2mm}\\ 
\textit{There is no reason to suppose that violent video games cause harm to children.}
\\\vspace{-3mm}\\ 
\textbf{Annotation:} Reject
\\\vspace{-3mm}\\ 
\textbf{Reason:} The candidate states a claim. It does not offer any additional information to convince the reader that this claim is true.
\\ \\ 
\underline{Example 4}
\\\vspace{-2mm}\\ 
\textit{The American Psychological Association argues that violent video-game play leads to increased moral sensitivity.}
\\\vspace{-3mm}\\ 
\textbf{Annotation:} Accept - Con
\\\vspace{-3mm}\\ 
\textbf{Reason:} The candidate states a claim, but the fact that it is raised by an authority figure (organization or human) turns it into a valuable evidence.
\\ \\ 
\underline{Example 5}
\\\vspace{-2mm}\\ 
\textit{Kennelly said there is no scientific evidence that violent video games cause “serious harm” in kids such as heightened aggression that would require protection of the law.}
\\\vspace{-3mm}\\ 
\textbf{Annotation:} Accept - Con
\\\vspace{-3mm}\\ 
\textbf{Note:}  If you are not certain whether the speaker is an authority figure or not, you should typically give him/her the benefit of the doubt and consider them as such (in this case the speaker is Matthew F. Kennelly, a United States District Judge). However,  if the candidate states a claim and the speaker is only mentioned by he/she/they you should reject it.
\\ \\
\underline{Example 6}
\\\vspace{-2mm}\\ 
\textit{The issue as “Psychological research confirms that violent video games can increase children's aggression.”}
\\\vspace{-3mm}\\ 
\textbf{Annotation:} Reject
\\\vspace{-3mm}\\ 
\textbf{Reason:} The candidate does not represent a coherent, stand-alone statement.
\\ \\
\underline{Example 7}
\\\vspace{-2mm}\\ 
\textit{Some studies have clearly demonstrated that video game violence is leading to serious aggressive behaviour in real life, although other studies have shown the opposite.}
\\\vspace{-3mm}\\ 
\textbf{Annotation:} Reject
\\\vspace{-3mm}\\ 
\textbf{Reason:} The pro/con stance of the candidate towards the topic is unclear, since the end of the text contradicts its beginning.
\\ \\ 
\underline{Example 8}
\\\vspace{-2mm}\\ 
\textit{The Entertainment Software Association reports that 17\% of violent video game players are boys under the age of eighteen.}
\\\vspace{-3mm}\\ 
\textbf{Annotation:} Reject
\\\vspace{-3mm}\\ 
\textbf{Reason:} The candidate states a fact with no clear pro/con stance towards the topic.
\\ \\
\underline{Example 9}
\\\vspace{-2mm}\\ 
\textit{Studies show that watching violent movies increases aggression amongst youth.}
\\\vspace{-3mm}\\ 
\textbf{Annotation:} Reject
\\\vspace{-3mm}\\ 
\textbf{Reason:} The candidate is not related to the topic as it discusses violent movies and not violent video games.
\\ \\
\underline{Example 10}
\\\vspace{-2mm}\\ 
\textit{Another 2001 meta-analyses and a more recent 2009 study focusing specifically on serious aggressive behavior concluded that video game violence is not related to serious aggressive behavior in real life.}
\\\vspace{-3mm}\\ 
\textbf{Annotation:} Accept - Con
\\\vspace{-3mm}\\ 
\textbf{Note:}  Even though the candidate’s first word better be omitted to make it a stand-alone statement, this is a minor change which is acceptable.
\\ \\
\underline{Example 11}
\\\vspace{-2mm}\\ 
\textit{Limiting the sale of violent video games will cause 15,000 people to lose their jobs.}
\\\vspace{-3mm}\\ 
\textbf{Annotation:} Accept - Con
\\\vspace{-3mm}\\ 
\textbf{Note:}  The candidate presents a specific numeric piece of information that clearly contest the topic. You are not expected to fact check the provided piece of information, don’t reject such a candidate just because you are not sure that the provided piece of information is true.